\documentclass[10pt,journal,compsoc]{IEEEtran}
\ifCLASSOPTIONcompsoc
  \usepackage[nocompress]{cite}
\else
  \usepackage{cite}
\fi
\ifCLASSINFOpdf
\else
\fi
\usepackage{graphicx}
\usepackage[skip=2pt,font=small]{caption}
\usepackage{amsmath}
\usepackage{amssymb}
\usepackage{booktabs}
\usepackage{multirow}
\usepackage{bm}
\usepackage{dsfont}
\usepackage{subcaption}

\usepackage{color}
\definecolor{feedforwarding}{RGB}{31,119,180}
\definecolor{backpropagation}{RGB}{214,39,40}

\usepackage{pifont}
\newcommand{\cmark}{\ding{51}}%

\newcommand{\vect}[1]{\boldsymbol{#1}}
\newcommand{\norm}[1]{\left\lVert#1\right\rVert}

\usepackage[ruled,vlined,linesnumbered]{algorithm2e}

\usepackage{enumitem}

\begin{document}
%
\title{
Micro-Batch Training with Batch-Channel Normalization and Weight Standardization
}
%
%
%
%

\author{
        Siyuan Qiao,~Huiyu Wang,~Chenxi Liu,~Wei Shen,~and Alan Yuille,~\textit{Fellow, IEEE}
\IEEEcompsocitemizethanks{\IEEEcompsocthanksitem All authors are with the Department
of Computer Science, Johns Hopkins University, Baltimore,
MD, 21218.
E-mail: \{siyuan.qiao, hwang157, cxliu\}@jhu.edu
\{shenwei1231, alan.l.yuille\}@gmail.com

\IEEEcompsocthanksitem Corresponding author: W. Shen

}
\thanks{Manuscript received April 19, 2005; revised August 26, 2015.}}

%
%

\markboth{Journal of \LaTeX\ Class Files,~Vol.~14, No.~8, August~2015}%
{Shell \MakeLowercase{\textit{et al.}}: Bare Demo of IEEEtran.cls for Computer Society Journals}
%



\IEEEtitleabstractindextext{%
\begin{abstract}

Batch Normalization (BN) has become an out-of-box technique to improve deep network training.
However, its effectiveness is limited for  
micro-batch training, \emph{i.e.}, each GPU typically has only 1-2 images for training, which is inevitable for many computer vision tasks, \textit{e.g.}, object detection and semantic segmentation, constrained by memory consumption. To address this issue, we propose Weight Standardization (WS) and Batch-Channel Normalization (BCN) to bring two success factors of BN into micro-batch training: 1) the smoothing effects on the loss landscape and 2) the ability to avoid harmful elimination singularities along the training trajectory.
WS standardizes the weights in convolutional layers to smooth the loss landscape by reducing the Lipschitz constants of the loss and the gradients; BCN combines batch and channel normalizations and leverages estimated statistics of the activations in convolutional layers to keep networks away from elimination singularities. 
We validate WS and BCN on comprehensive computer vision tasks, including image classification, object detection, instance segmentation, video recognition and semantic segmentation.
All experimental results consistently show that WS and BCN improve micro-batch training significantly. Moreover, using WS and BCN with micro-batch training is even able to match or outperform the performances of BN with large-batch training.

\end{abstract}

}

\maketitle

\IEEEdisplaynontitleabstractindextext

%
\IEEEpeerreviewmaketitle

\begin{figure}[t]
    \centering
    \includegraphics[width=\linewidth]{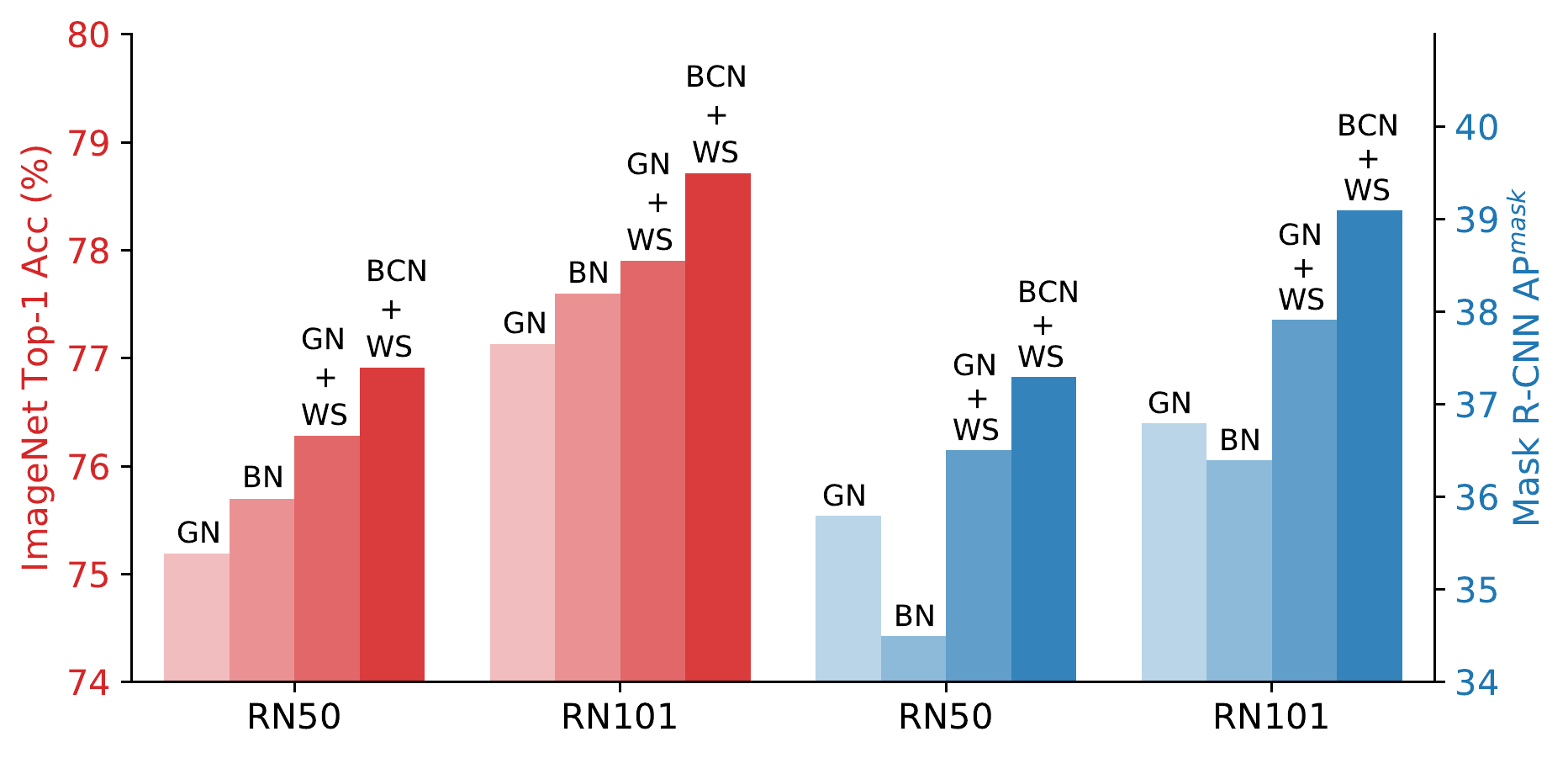}
    \caption{Comparing BN~\cite{batchnorm}, GN~\cite{groupnorm}, our WS used with GN, and WS used with BCN on ImageNet and COCO.
    On ImageNet, BN and BCN+WS are trained with large batch sizes while GN and GN+WS are trained with 1 image/GPU.
    On COCO, BN is frozen for micro-batch training, and BCN uses its micro-batch implementation.
    GN+WS outperforms both BN and GN comfortably and BCN+WS further improves the performances.}
    \label{fig:front}
\end{figure}

\IEEEraisesectionheading{\section{Introduction}\label{sec:intro}}
Deep learning has advanced the state-of-the-arts in many vision tasks~\cite{deeplab,resnet}.
Many deep networks use Batch Normalization (BN)~\cite{batchnorm} in their architectures because BN in most cases is able to accelerate training and help the models to converge to better solutions.
BN stabilizes the training by controlling the first two moments of the distributions of the layer outputs in each mini-batch during training and is especially helpful for training very deep networks that have hundreds of layers~\cite{resnetv2,densenet}.
Despite its practical success, BN has a shortcoming that it works well only when the batch size is sufficiently large, which prohibits it from being used in micro-batch training. Micro-batch training, \emph{i.e.}, the batch size is small, \emph{e.g.}, 1 or 2, is inevitable for many computer vision tasks, such as object detection and semantic segmentation, due to limited GPU memory. This shortcoming draws a lot of attentions from researchers, which urges them to design specific normalization methods for micro-batch training, such as Group Normalization (GN)~\cite{groupnorm} and Layer Normalization (LN)~\cite{layernorm}, but they have difficulty matching the performances of BN in large-batch training (Fig.~\ref{fig:front}).

In this paper, our goal is to bring the success factors of BN into micro-batch training but without relying on large batch sizes during training. This requires good understandings of the reasons of BN's success, among which we focus on two factors:
\begin{enumerate}[wide]
    \item {\bf BN's smoothing effects:}
        \cite{whybnworks} proves that BN makes the landscape of the corresponding optimization problem significantly smoother, thus is able to stabilize the training process and accelerate the convergence speed of training deep neural networks.
    \item {\bf BN avoids elimination singularities:}
        Elimination singularities refer to the points along the training trajectory where neurons in the networks get eliminated.
        Eliminable neurons waste computations and decrease the effective model complexity.
        Getting closer to them will harm the training speed and the final performances.
        By forcing each neuron to have zero mean and unit variance, BN keeps the networks at far distances from elimination singularities caused by non-linear activation functions.
\end{enumerate}

We find that these two success factors are not properly addressed by some methods specifically designed for micro-batch training.
For example, channel-based normalizations, \textit{e.g.}, Layer Normalization (LN)~\cite{layernorm} and Group Normalization (GN)~\cite{groupnorm}, are \textbf{unable} to guarantee far distances from elimination singularities.
This might be the reason for their inferior performance compared with BN in large-batch training.
To bring the above two success factors into micro-batch training, we propose Weight Standardization (WS) and Batch-Channel Normalization (BCN) to improve network training.
WS standardizes the weights in convolutional layers, \emph{i.e.}, making the weights have zero mean and unit variance.
BCN uses estimated means and variances of the activations in convolutional layers by combining batch and channel normalization.
WS and BCN are able to run in both large-batch and micro-batch settings and accelerate the training and improve the performances.

We study WS and BCN from both theoretical and experimental viewpoints.
The highlights of the results are:
\begin{enumerate}
    \item Theoretically, we prove that WS reduces the Lipschitz constants of the loss and the gradients.
    Hence, WS smooths loss landscape and improves training.
    \item We empirically show that WS and BCN are able to push the models away from the elimination singularities.
    \item Experiments show that on tasks where large-batches are available (\textit{e.g.} ImageNet~\cite{ILSVRC15}), GN~\cite{groupnorm} + WS with batch size 1 is able to match or outperform the performances of BN with large batch sizes (Fig.~\ref{fig:front}).
    \item For tasks where only micro-batch training is available (\textit{e.g.} COCO~\cite{coco}), GN + WS will significantly improve the performances (Fig.~\ref{fig:front}).
    \item Replacing GN with BCN further improves the results in both large-batch and micro-batch training settings.
\end{enumerate}

To show that our WS and BCN are applicable to many vision tasks, we conduct comprehensive experiments, including image classification on CIFAR-10/100~\cite{cifar} and ImageNet dataset~\cite{ILSVRC15}, object detection and instance segmentation on MS COCO dataset~\cite{coco},
video recognition on Something-SomethingV1 dataset~\cite{something},
and semantic image segmentation on PASCAL VOC~\cite{pascal}.
The experimental results show that our WS and BCN are able to accelerate training and improve performances.

\section{Related Work}\label{sec:related}
Deep neural networks advance state-of-the-arts in many computer vision tasks~\cite{deeplab,densenet,alexnet,fcnn,fewshot,qiao2018deep,qiu2017unrealcv,vggnet,sort,wang2018multi,yang2018knowledge,zhang2018single}.
But deep networks are hard to train.
To speed up training, proper model initialization strategies are widely used as well as data normalization based on the assumption of the data distribution~\cite{glorot2010understanding,he2015delving}.
On top of data normalization and model initialization, Batch Normalization~\cite{batchnorm} is proposed to ensure certain distributions so that the normalization effects will not fade away during training.
By performing normalization along the batch dimension, Batch Normalization achieves state-of-the-art performances in many tasks in addition to accelerating the training process.
When the batch size decreases, however, the performances of Batch Normalization drop dramatically since the batch statistics are not representative enough of the dataset statistics.
Unlike Batch Normalization that works on the batch dimension, Layer Normalization~\cite{layernorm} normalizes data on the channel dimension, Instance Normalization~\cite{instnorm} does Batch Normalization for each sample individually.
Group Normalization~\cite{groupnorm} also normalizes features on the channel dimension, but it finds a better middle point between Layer Normalization and Instance Normalization.

Batch Normalization, Layer Normalization, Group Normalization, and Instance Normalization are all activation-based normalization methods.
Besides them, there are also weight-based normalization methods, such as Weight Normalization~\cite{weightnorm} and Centered Weight Normalization~\cite{huang2017centered}.
Weight Normalization decouples the length and the direction of the weights, and Centered Weight Normalization also centers the weights to have zero mean.
Weight Standardization is similar, but removes the learnable weight length.
Instead, the weights are standardized to have zero mean and unit variance, and then directly sent to the convolution operations.
When used with GN, it narrows the performance gap between BN and GN.

In this paper, we study normalization from the perspective of elimination singularity~\cite{orhan2017skip,wei2008dynamics} and smoothness~\cite{whybnworks}.
There are also other perspectives to understand normalization methods.
For example, from training robustness, BN is able to make optimization trajectories more robust to parameter initialization~\cite{im2016empirical}.
\cite{whybnworks} shows that normalizations are able to reduce the Lipschitz constants of the loss and the gradients, thus the training becomes easier and faster.
From the angle of model generalization, \cite{morcos2018importance} shows that Batch Normalization relies less on single directions of activations, thus has better generalization properties, and \cite{luo2018towards} studies the regularization effects of Batch Normalization.
\cite{kohler2018towards} also explores length-direction decoupling in BN and WN~\cite{weightnorm}.
Other work also approaches normalizations from the gradient explosion issues~\cite{DBLP:journals/corr/abs-1902-08129} and learning rate tuning~\cite{DBLP:journals/corr/abs-1812-03981}.
Our WS is also related to converting constrained optimization to unconstrained optimization~\cite{absil2009optimization,cho2017riemannian}.

Our BCN uses Batch Normalization and Group Normalization at the same time for one layer.
Some previous work also uses multiple normalizations or a combined version of normalizations for one layer.
For example, SN~\cite{switchnorm} computes BN, IN, and LN at the same time and uses AutoML~\cite{pnas} to determine how to combine them.
SSN~\cite{shao2019ssn} uses SparseMax to get sparse SN.
DN~\cite{dynamicnorm} proposes a more flexible form to represent normalizations and finds better normalizations.
Unlike them, our method is based on analysis and theoretical understandings instead of searching solutions through AutoML, and our normalizations are used together as a composite function rather than linearly adding up the normalization effects in a flat way.

\section{Lipschitz Smoothness and Elimination Singularities}\label{sec:pre}
We first describe Lipschitz Smoothness and Elimination Singularities to provide the background of our analyses.

\subsection{Lipschitz Smoothness}
A function $f: A\rightarrow \mathbb{R}^m$, $A\in\mathbb{R}^n$ is $L$-Lipschitz \cite{basiccourse} if
\begin{equation}
    \forall a, b \in A: ~~ ||f(a) - f(b)|| \leq L || a - b ||.
\end{equation}
A continuously differentiable function $f$ is $\beta$-smooth if the gradient $\nabla f$ is
$\beta$-Lipschitz, \textit{i.e.},
\begin{equation}
    \forall a, b \in A: ~~ ||\nabla f(a) - \nabla f(b) || \leq \beta || a - b ||.
\end{equation}

Many results show that training smooth functions using gradient descent algorithms is faster than training non-smooth functions~\cite{bubeck2015convex}.
Intuitively, gradient descent based training algorithms can be unstable due to exploding or vanishing gradients.
As a result, they are sensitive to the selections of the learning rate and initialization if the loss landscape is not smooth.
Using an algorithm (\textit{e.g.} WS) that smooths the loss landscape will make the gradients more reliable and predictive; thus, larger steps can be taken and the training will be accelerated.

\subsection{Elimination singularities}
Deep neural networks are hard to train partly due to the singularities caused by the non-identifiability of the model~\cite{wei2008dynamics}.
These singularities include overlap singularities, linear dependence singularities, elimination singularities, \textit{etc}.
Degenerate manifolds in the loss landscape will be caused by these singularities, getting closer to which will slow down learning and impact model performances~\cite{orhan2017skip}.
In this paper, we focus on elimination singularities, which correspond to the points on the training trajectory where neurons in the model become constantly deactivated.

The original definition of elimination singularities is based on weights~\cite{wei2008dynamics}: if we use $\vect{\omega}_c$ to denote the weights that take the channel $c$ as input,
then an elimination singularity is encountered when $\vect{\omega}_c=\bm{0}$.
However, this definition is not suitable for real-world deep network training as most of $\vect{\omega}_c$ will not be close to $0$.
For example, in a ResNet-50~\cite{resnet} well-trained on ImageNet~\cite{ILSVRC15}, $
    \frac{1}{L}\sum_{l}\frac{\text{min}_{c\in l}{ |\vect{\omega}_c|_1 }}{\text{avg}_{c\in l}{ |\vect{\omega}_c|_1 }} = 0.55
$, where $L$ is the number of all the layers $l$ in the network.
Note that weight decay is already used in training this network to encourage weight sparsity.
In other words, defining elimination singularities based on weights is not proper for networks trained in real-world settings.

In this paper, we consider elimination singularities for networks that use ReLU as their activation functions.
We focus on a basic building element that is widely used in neural networks: a convolutional layer followed by a normalization method (\textit{e.g.} BN, LN) and ReLU~\cite{relu}, \textit{i.e.},
\begin{equation}
   \vect{X^{\text{out}}} = \text{ReLU}(\text{Norm}(\text{Conv}(\vect{X^{\text{in}}}))).
\end{equation}
When ReLU is used, $\vect{\omega}_c=\bm{0}$ is no longer necessary for a neuron to be eliminatable.
This is because ReLU sets any values below $0$ to $0$; thus a neuron is constantly deactivated if its maximum value after the normalization layer is below $0$.
Their gradients will also be $0$ because of ReLU, making them hard to revive; hence, a singularity is created.

\begin{figure}[t]
    \centering
    \includegraphics[width=\linewidth]{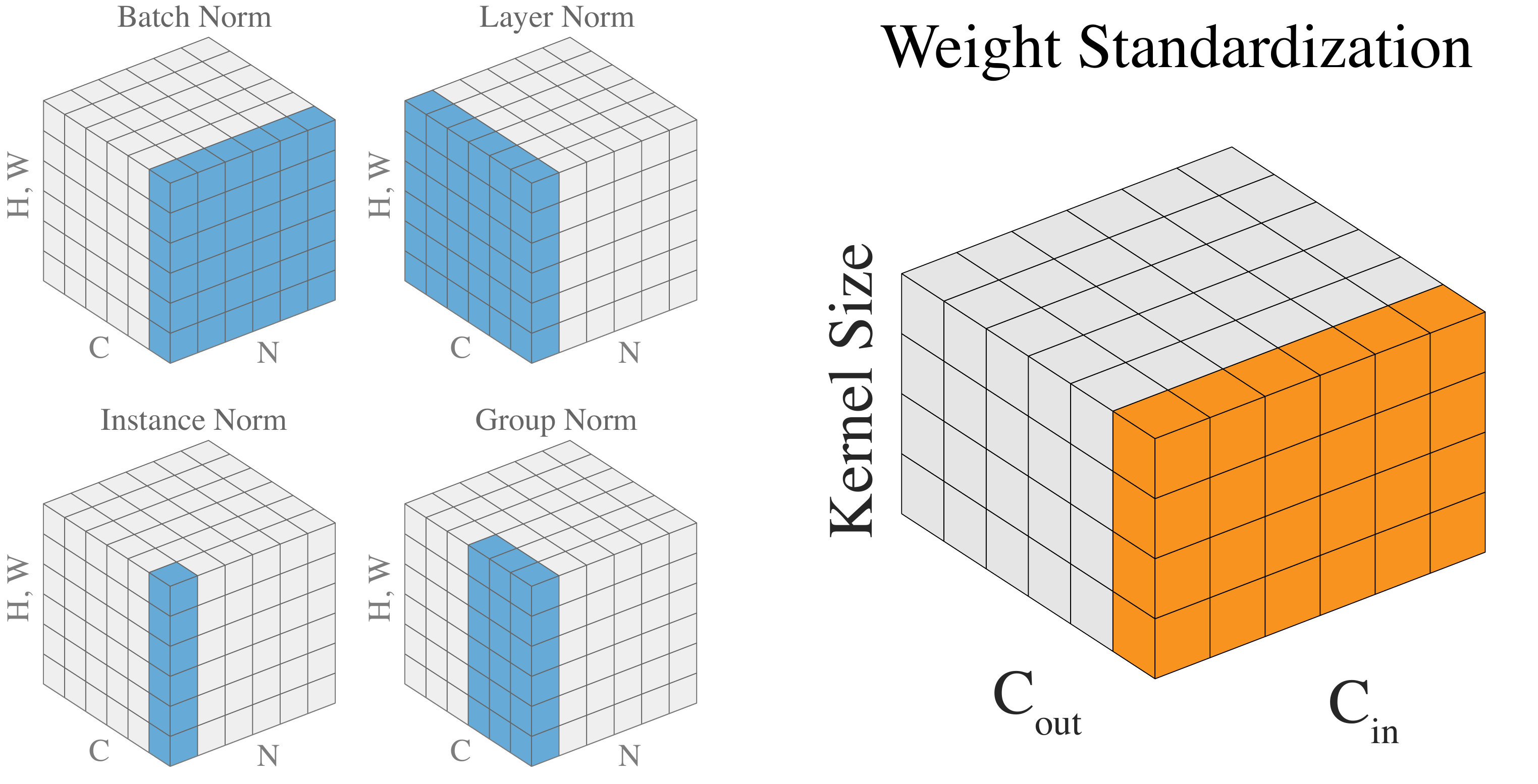}
    \caption{Comparing normalization methods on activations (blue) and Weight Standardization (orange).}
    \label{fig:all}
\end{figure}

\section{Weight Standardization}\label{sec:ws}
In this section, we introduce Weight Standardization, which is inspired by BN.
It has been demonstrated that BN influences network training in a fundamental way: it makes the landscape of
the optimization problem significantly smoother~\cite{whybnworks}.
Specifically, \cite{whybnworks} shows that BN reduces the Lipschitz constants of the loss function, and makes the gradients more Lipschitz, too,
\textit{i.e.}, the loss will have a better $\beta$-smoothness~\cite{basiccourse}.

We notice that BN considers the Lipschitz constants with respect to \emph{activations}, not the \emph{weights} that the optimizer is directly optimizing.
Therefore, we argue that we can also standardize the \emph{weights} in the convolutional layers to further smooth the landscape.
By doing so, we do not have to worry about transferring smoothing effects from activations to weights; moreover, the smoothing effects on activations and weights are also additive.
Based on these motivations, we propose Weight Standardization.

\subsection{Weight Standardization}

Here, we show the detailed modeling of Weight Standardization (WS) (Fig.~\ref{fig:all}).
Consider a standard convolutional layer with its bias term set to 0:
\begin{equation}
    \vect{y} = \hat{\vect{W}}*\vect{x},
\end{equation}
where $\hat{\vect{W}}\in\mathbb{R}^{O\times I}$ denotes the weights in the layer and $*$ denotes the convolution operation.
For $\hat{\vect{W}}\in\mathbb{R}^{O\times I}$, $O$ is the number of the output channels, $I$ corresponds to the number of input channels within the kernel region of each output channel.
Taking Fig.~\ref{fig:all} as an example, $O=C_{\text{out}}$ and $I=C_{\text{in}}\times\text{Kernel\_Size}$.
In Weight Standardization, instead of directly optimizing the loss $\mathcal{L}$ on the original weights $\hat{\vect{W}}$, we reparameterize the weights $\hat{\vect{W}}$ as a function of $\vect{W}$, \textit{i.e.}, $\hat{\vect{W}}=\text{WS}(\vect{W})$, and optimize the loss $\mathcal{L}$ on $\vect{W}$ by SGD:
\begin{align}
    \hat{\vect{W}} &= \Big[ \hat{\vect{W}}_{i,j}~\big|~ \hat{\vect{W}}_{i,j} = \dfrac{\vect{W}_{i,j} - \mu_{\vect{W}_{i,\cdot}}}{\sigma_{\vect{W}_{i,\cdot}}}\Big]\label{eq:6},\\
    \vect{y} &= \hat{\vect{W}}*\vect{x}\label{eq:7},
\end{align}
where
\begin{align}\label{eq:epsilon}
    \mu_{\vect{W}_{i,\cdot}} = \dfrac{1}{I}\sum_{j=1}^{I}\vect{W}_{i, j},~~\sigma_{\vect{W}_{i,\cdot}}=\sqrt{\dfrac{1}{I}\sum_{j=1}^I\vect{W}_{i,j}^2 - \mu_{\vect{W}_{i,\cdot}}^2 + \epsilon}.
\end{align}

Similar to BN, WS controls the first and second moments of the weights of each output channel individually in convolutional layers.
Note that many initialization methods also initialize the weights in some similar ways.
Different from those methods, WS standardizes the weights in a differentiable way which aims to normalize gradients during back-propagation.
Note that we do not have any affine transformation on $\hat{\vect{W}}$.
This is because we assume that normalization layers such as BN or GN will normalize this convolutional layer again, and having affine transformation will confuse and slow down training.
In the following, we first discuss the normalization effects of WS to the gradients.

\subsection{Comparing WS with WN and CWN}
Weight Normalization (WN)~\cite{weightnorm} and Centered Weight Normalization (CWN)~\cite{huang2017centered} also normalize weights to speed up deep network training.
Weight Normalization reparameterizes weights by separating the direction $\frac{W}{\norm{W}}$ and length $g$:
\begin{equation}
    \hat{\vect{W}} = g\frac{\vect{W}}{\norm{\vect{W}}}.
\end{equation}
WN is able to train good models on many tasks.
But as shown in \cite{gitman2017comparison}, WN has difficulty matching the performances of models trained with BN on large-scale datasets.
Later, CWN adds a centering operation for WN, \textit{i.e.},
\begin{equation}\label{eq:cwn}
    \hat{W} = g\frac{\vect{W}-\overline{\vect{W}}}{\norm{\vect{W} - \overline{\vect{W}}}}.
\end{equation}
To compare with WN and CWN, we consider the weights for only one of the output channel and reformulate the corresponding weights output by WS in Eq.~\ref{eq:6} as
\begin{equation}
    \hat{W} = \dfrac{\vect{W}-\overline{\vect{W}}}{\sqrt{\overline{\vect{W}^2} - {\overline{\vect{W}}}^2}},
\end{equation}
which removes the learnable length $g$ from Eq.~\ref{eq:cwn} and divides the weights with their standard deviation instead.
Experiments in Sec.~\ref{sec:exp} show that WS outperforms WN and CWN on large-scale tasks~\cite{ILSVRC15}.

\section{The smoothing effects of WS}

In this section, we discuss the smoothing effects of WS.
Sec.~\ref{sec:wng} shows that WS normalizes the gradients.
This normalization effect on gradients lowers the Lipschitz constants of the loss and the gradients as will be shown in Sec.~\ref{sec:wsl}, where Sec.~\ref{sec:eowot} discusses the effects on the loss and Sec.~\ref{sec:eowotl} discusses the effects on the gradients.

\subsection{WS normalizes gradients}\label{sec:wng}

\begin{figure}[t]
    \centering
    \includegraphics[width=\linewidth]{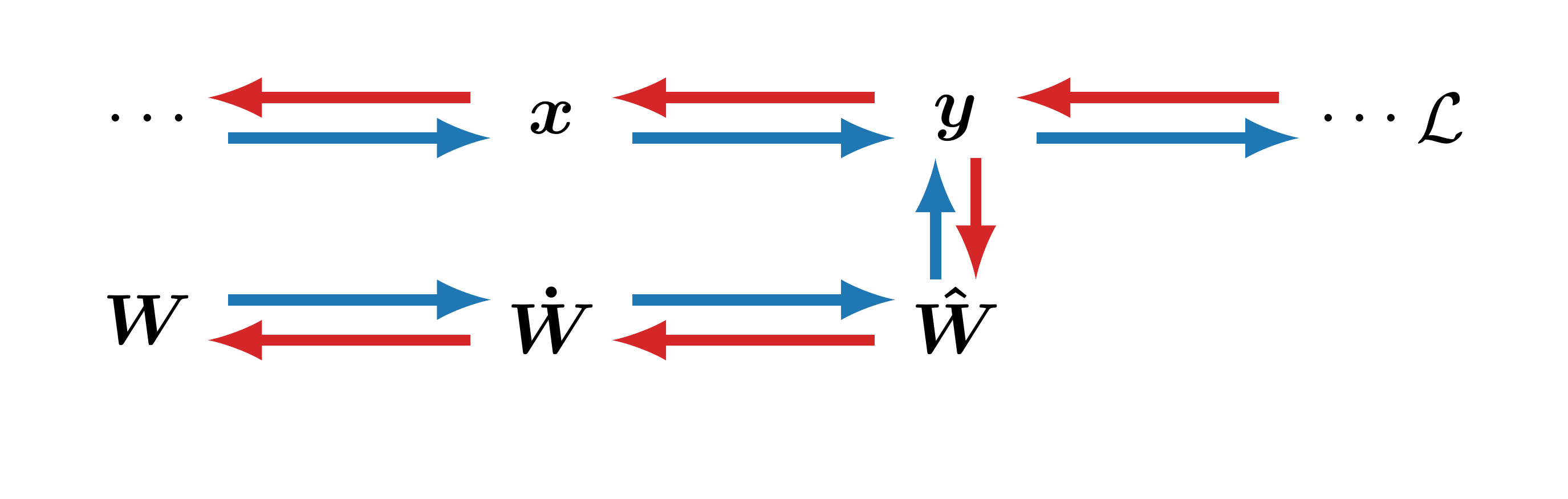}
    \caption{Computation graph for WS in {\color{feedforwarding}feed-forwarding} and {\color{backpropagation}back-propagation}.
    $\vect{W}$, $\dot{\vect{W}}$ and $\hat{\vect{W}}$ are defined in Eq.~\ref{eq:9}, \ref{eq:10} and \ref{eq:11}.}
    \label{fig:ws_eqn}
\end{figure}

For convenience, we set $\epsilon=0$ (in Eq.~\ref{eq:epsilon}).
We first focus on one output channel $c$.
Let $\vect{y}_c\in\mathbb{R}^{b}$ be all the outputs of channel $c$ during one pass of feedforwarding and back-propagation, and $\vect{x}_c\in\mathbb{R}^{b,I}$ be the corresponding inputs.
Then, we can rewrite Eq.~\ref{eq:6} and ~\ref{eq:7} as
\begin{align}
    \dot{\vect{W}}_{c,\cdot} &= \vect{W}_{c,\cdot} - \dfrac{1}{I}\mathbf{1}\langle\mathbf{1}, \vect{W}_{c,\cdot}\rangle\label{eq:9},\\
    \hat{\vect{W}}_{c,\cdot} &= \dot{\vect{W}}_{c,\cdot} / \Big( \sqrt{\dfrac{1}{I}\langle\mathbf{1},  \dot{\vect{W}}^{\circ 2}_{c,\cdot}\rangle}\Big)\label{eq:10},\\
    \vect{y}_c &=  \vect{x}_c \hat{\vect{W}}_{c,\cdot}\label{eq:11},
\end{align}
where $\langle~,~\rangle$ denotes dot product and $^{\circ 2}$ denotes Hadamard power.
Then, the gradients are
\begin{align}
    &\nabla_{\dot{\vect{W}}_{c,\cdot}}\mathcal{L} = \dfrac{1}{\sigma_{\vect{W}_{c,\cdot}}} \Big(  \nabla_{\hat{\vect{W}}_{c,\cdot}}\mathcal{L} - \dfrac{1}{I} \langle \hat{\vect{W}}_{c,\cdot}, \nabla_{\hat{\vect{W}}_{c,\cdot}}\mathcal{L} \rangle \hat{\vect{W}}_{c,\cdot} \Big),  \label{eq:12}\\
    &\nabla_{\vect{W}_{c,\cdot}}\mathcal{L} = \nabla_{\dot{\vect{W}}_{c,\cdot}}\mathcal{L} - \dfrac{1}{I}\mathbf{1}\langle\mathbf{1}, \nabla_{\dot{\vect{W}}_{c,\cdot}}\mathcal{L}\rangle\label{eq:13}.
\end{align}
Fig.~\ref{fig:ws_eqn} shows the computation graph.
Based on the equations, we observe that different from the original gradients $\nabla_{\hat{\vect{W}}_{c,\cdot}}\mathcal{L}$ which is back-propagated through Eq.~\ref{eq:11}, the gradients are normalized by Eq.~\ref{eq:12} \& \ref{eq:13}.

In Eq.~\ref{eq:12}, to compute $\nabla_{\dot{\vect{W}}_{c,\cdot}}\mathcal{L}$, $\nabla_{\hat{\vect{W}}_{c,\cdot}}\mathcal{L}$ is first subtracted by a weighted average of $\nabla_{\hat{\vect{W}}_{c,\cdot}}\mathcal{L}$ and then divided by $\sigma_{\hat{\vect{W}}_{c,\cdot}}$.
Note that when BN is used to normalize this convolutional layer, as BN will compute again the scaling factor $\sigma_{\vect{u}}$, the effects of dividing the gradients by $\sigma_{\hat{\vect{W}}_{c,\cdot}}$ will be canceled in both feedforwarding and back-propagation.
As for the additive term, its effect will depend on the statistics of $\nabla_{\hat{\vect{W}}_{c,\cdot}}\mathcal{L}$ and ${\hat{\vect{W}}}_{c,\cdot}$.
We will later show that this term will reduce the gradient norm regardless of the statistics.
As for Eq.~\ref{eq:13}, it zero-centers the gradients from $\dot{\vect{W}}_{c,\cdot}$.
When the mean gradient is large, zero-centering will significantly affect the gradients passed to $\vect{W}_{c,\cdot}$.


\begin{figure*}
    \centering
    \includegraphics[width=\linewidth]{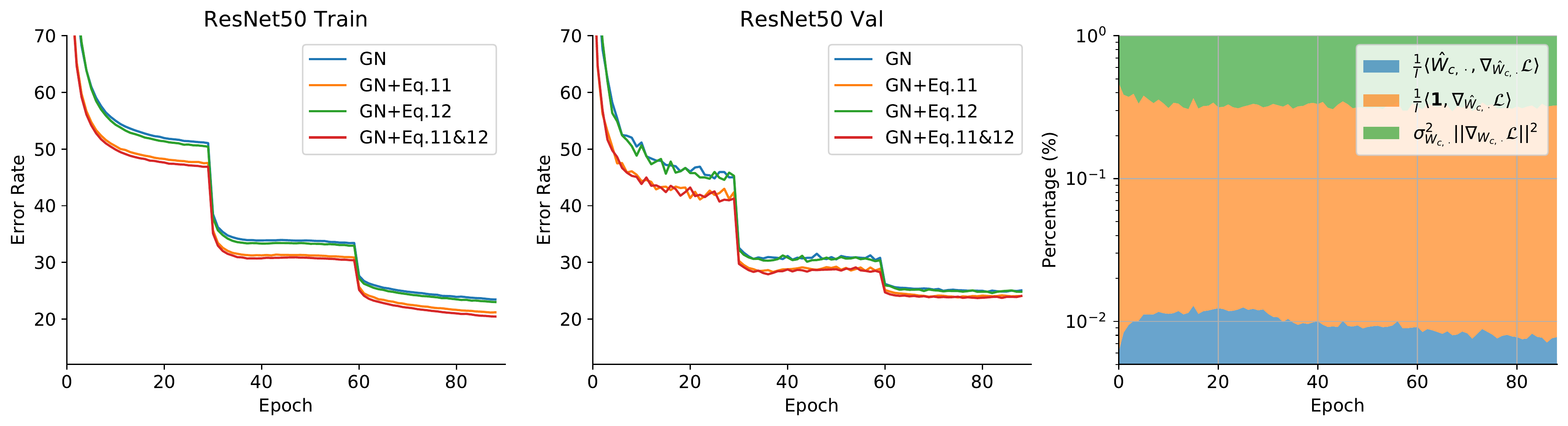}
    \caption{Training ResNet-50 on ImageNet with GN, Eq.~\ref{eq:9} and \ref{eq:10}.
    The left and the middle figures show the training dynamics.
    The right figure shows the reduction percentages on the Lipschitz constant.
    Note that the y-axis of the right figure is in {\bf log} scale.
    }
    \label{fig:mean_div}
\end{figure*}

\subsection{WS smooths landscape}\label{sec:wsl}
We will show that WS is able to make the loss landscape smoother.
Specifically, we show that optimizing $\mathcal{L}$ on $\vect{W}$ has smaller Lipschitz constants on both the loss and the gradients than optimizing $\mathcal{L}$ on $\hat{\vect{W}}$.
Lipschitz constant of a function $f$ is the value of $L$ if $f$ satisfies
$|f(x_1) - f(x_2)| \leq L \lVert x_1 - x_2\rVert,~\forall x_1,x_2$.
For the loss and gradients, $f$ will be $\mathcal{L}$ and $\nabla_{\vect{W}}\mathcal{L}$, and $x$ will be $\vect{W}$.
Smaller Lipschitz constants on the loss and gradients mean that the changes of the loss and the gradients during training will be bounded more.
They will provide more confidence when the optimizer takes a big step in the gradient direction as the gradient direction will vary less within the range of the step.
In other words, the optimizer can take longer steps without worrying about sudden changes of the loss landscape and gradients.
Therefore, WS is able to accelerate training.

\subsubsection{Effects of WS on the Lipschitz constant of the loss}\label{sec:eowot}
Here, we show that both Eq.~\ref{eq:12} and Eq.~\ref{eq:13} are able to reduce the Lipschitz constant of the loss.
We first study Eq.~\ref{eq:12}:
\begin{align}
\begin{split}
    \big\lVert \nabla_{\dot{{\vect{W}}}_{c,\cdot}}\mathcal{L} \big\rVert^2=\dfrac{1}{\sigma_{{W}_{c,\cdot}}^2}\Big( \big\lVert \nabla_{\hat{\vect{W}}_{c,\cdot}}\mathcal{L} \big\rVert^2 + \\ \dfrac{1}{I^2}\big\langle \hat{\vect{W}}_{c,\cdot}, \nabla_{\hat{\vect{W}}_{c,\cdot}}\mathcal{L} \big\rangle^2\big( \langle \hat{\vect{W}}_{c,\cdot}, \hat{\vect{W}}_{c,\cdot} \rangle - 2I \big)\Big).
\end{split}
\end{align}
By Eq.~\ref{eq:10}, we know that $\lVert\hat{\vect{W}}_{c,\cdot}\rVert^2=I$.
Then,
\begin{align}
\begin{split}
    \big\lVert \nabla_{\dot{{\vect{W}}}_{c,\cdot}}\mathcal{L} \big\rVert^2=\dfrac{1}{\sigma_{{W}_{c,\cdot}}^2}\Big( \big\lVert \nabla_{\hat{\vect{W}}_{c,\cdot}}\mathcal{L} \big\rVert^2 - \\\dfrac{1}{I}\big\langle \hat{\vect{W}}_{c,\cdot}, \nabla_{\hat{\vect{W}}_{c,\cdot}}\mathcal{L} \big\rangle^2\Big).
\end{split}
\end{align}
Since we assume that this convolutional layer is followed by a normalization layer such as BN or GN, the effect of $1/\sigma_{{W}_{c,\cdot}}^2$ will be canceled.
Therefore, the real effect on the gradient norm is the reduction $\dfrac{1}{I}\big\langle \hat{\vect{W}}_{c,\cdot}, \nabla_{\hat{\vect{W}}_{c,\cdot}}\mathcal{L} \big\rangle^2$, which reduces the Lipschitz constant of the loss.

Next, we study the effect of Eq.~\ref{eq:13}.
By definition,
\begin{equation}
    \big\lVert \nabla_{\vect{W}_{c,\cdot}}\mathcal{L} \big\rVert^2 = \big\lVert \nabla_{\dot{{\vect{W}}}_{c,\cdot}}\mathcal{L} \big\rVert^2 - \dfrac{1}{I} \langle\mathbf{1}, \nabla_{\dot{{\vect{W}}}_{c,\cdot}}\mathcal{L}\rangle^2.
\end{equation}
By Eq.~\ref{eq:12}, we rewrite the second term:
\begin{align}
\begin{split}
    \dfrac{1}{I} \langle\mathbf{1}, \nabla_{\dot{{\vect{W}}}_{c,\cdot}}\mathcal{L}\rangle^2 = \dfrac{1}{I\cdot\sigma^2_{W_{c,\cdot}}} \Big( \langle\mathbf{1}, \nabla_{\hat{{\vect{W}}}_{c,\cdot}}\mathcal{L}\rangle \\
    - \dfrac{1}{I} \langle \hat{\vect{W}}_{c,\cdot}, \nabla_{\hat{\vect{W}}_{c,\cdot}}\mathcal{L} \rangle \cdot \langle \mathbf{1},~ \hat{\vect{W}}_{c,\cdot} \rangle\Big)^2.
\end{split}
\end{align}
Since $\langle \mathbf{1},~ \hat{\vect{W}}_{c,\cdot} \rangle=0$, we have
\begin{equation}
    \big\lVert \nabla_{\vect{W}_{c,\cdot}}\mathcal{L} \big\rVert^2 = \big\lVert \nabla_{\dot{{\vect{W}}}_{c,\cdot}}\mathcal{L} \big\rVert^2 - \dfrac{1}{I\cdot\sigma^2_{W_{c,\cdot}}} \langle\mathbf{1}, \nabla_{\hat{{\vect{W}}}_{c,\cdot}}\mathcal{L}\rangle^2.
\end{equation}

Summarizing the effects of Eq.~\ref{eq:12} and \ref{eq:13} on the Lipschitz constant of the loss: ignoring $1/\sigma_{{W}_{c,\cdot}}^2$, Eq.~\ref{eq:12} reduces it by $\frac{1}{I}\big\langle \hat{\vect{W}}_{c,\cdot}, \nabla_{\hat{\vect{W}}_{c,\cdot}}\mathcal{L} \big\rangle^2$, and Eq.~\ref{eq:13} reduces it by $\frac{1}{I} \langle\mathbf{1}, \nabla_{\hat{{\vect{W}}}_{c,\cdot}}\mathcal{L}\rangle^2$.

Although both Eq.~\ref{eq:12} and \ref{eq:13} reduce the Lipschitz constant, their real effects depend on the statistics of the weights and the gradients.
For example, the reduction effect of Eq.~\ref{eq:13} depends on the average gradients on $\hat{\vect{W}}$.
As for Eq.~\ref{eq:12}, note that $\langle \mathbf{1},~ \hat{\vect{W}}_{c,\cdot} \rangle=0$, its effect might be limited when $\hat{\vect{W}}_{c,\cdot}$ is evenly distributed around $0$.
To understand their real effects, we conduct a case study on ResNet-50 trained on ImageNet to see which one of Eq.~\ref{eq:9} and \ref{eq:10} has bigger effects or they contribute similarly to smoothing the landscape.

\subsubsection{Effects of WS on the Lipschitz constant of gradients}\label{sec:eowotl}
Before the Lipschitzness study on the gradients, we first show a case study where we train ResNet-50 models on ImageNet following the conventional training procedure~\cite{resnet}.
In total, we train four models, including ResNet-50 with GN, ResNet-50 with GN+Eq.~\ref{eq:9}, ResNet-50 with GN+Eq.~\ref{eq:10} and ResNet-50 with GN+Eq.~\ref{eq:9}\&\ref{eq:10}.
The training dynamics are shown in Fig.~\ref{fig:mean_div},
from which we observe that Eq.~\ref{eq:10} slightly improves the training speed and performances of models with or without Eq.~\ref{eq:9} while the major improvements are from Eq.~\ref{eq:9}.
This observation motivates us to study the real effects of Eq.~\ref{eq:9} and~\ref{eq:10} on the Lipschitz constant of the loss.
To investigate this, we take a look at the values of $\frac{1}{I}\big\langle \hat{\vect{W}}_{c,\cdot}, \nabla_{\hat{\vect{W}}_{c,\cdot}}\mathcal{L} \big\rangle^2$, and $\frac{1}{I} \langle\mathbf{1}, \nabla_{\hat{{\vect{W}}}_{c,\cdot}}\mathcal{L}\rangle^2$ during training.

To compute the two values above, we gather and save the intermediate gradients $\nabla_{\hat{\vect{W}}_{c,\cdot}}\mathcal{L}$, and the weights for the convolution $\hat{\vect{W}}_{c,\cdot}$.
In total, we train ResNet-50 with GN, Eq.~\ref{eq:9} and \ref{eq:10} for 90 epochs, and we save the gradients and the weights of the first training iteration of each epoch.
The right figure of Fig.~\ref{fig:mean_div} shows the average percentages of $\frac{1}{I}\big\langle \hat{\vect{W}}_{c,\cdot}, \nabla_{\hat{\vect{W}}_{c,\cdot}}\mathcal{L} \big\rangle^2$, $\frac{1}{I} \langle\mathbf{1}, \nabla_{\hat{{\vect{W}}}_{c,\cdot}}\mathcal{L}\rangle^2$, and $\sigma^2_{\vect{W}_{c,\cdot}}\lVert \nabla_{\vect{W}_{c,\cdot}}\mathcal{L} \rVert^2$.
From the right figure we can see that $\frac{1}{I}\big\langle \hat{\vect{W}}_{c,\cdot}, \nabla_{\hat{\vect{W}}_{c,\cdot}}\mathcal{L} \big\rangle^2$ is small compared with other two components ($<0.02$).
In other words, although Eq.~\ref{eq:10} decreases the gradient norm regardless of the statistics of the weights and gradients, its real effect is limited due to the distribution of $\hat{\vect{W}}_{c,\cdot}$ and $\nabla_{\hat{\vect{W}}_{c,\cdot}}\mathcal{L}$.
Nevertheless, from the left figures we can see that Eq.~\ref{eq:10} still improves the training.
Since Eq.~\ref{eq:10} requires very little computations, we will keep it in WS.

From the experiments above, we observe that the training speed boost is mainly due to Eq.~\ref{eq:9}.
As the effect of Eq.~\ref{eq:10} is limited, in this section, we only study the effect of Eq.~\ref{eq:9} on the Hessian of $\vect{W}_{c,\cdot}$ and $\dot{\vect{W}}_{c,\cdot}$.
Here, we will show that Eq.~\ref{eq:9} decreases the Frobenius norm of the Hessian matrix of the weights, \textit{i.e.}, $\lVert \nabla^2_{\vect{W}_{c,\cdot}}\mathcal{L} \rVert_{F} \leq \lVert \nabla^2_{\dot{\vect{W}}_{c,\cdot}}\mathcal{L} \rVert_{F}$.
With smaller Frobenius norm, the gradients of $\vect{W}_{c,\cdot}$ are more predictable, thus the loss is smoother and easier to optimize.

We use $\vect{H}$ and $\dot{\vect{H}}$ to denote the Hessian matrices of $\vect{W}_{c,\cdot}$ and $\dot{\vect{W}}_{c,\cdot}$, respectively, \textit{i.e.},
\begin{equation}
    \vect{H}_{i,j} = \dfrac{\partial^2\mathcal{L}}{\partial\vect{W}_{c,i}\partial\vect{W}_{c,j}}, ~~~\dot{\vect{H}}_{i,j} = \dfrac{\partial^2\mathcal{L}}{\partial\vect{\dot{W}}_{c,i}\partial\vect{\dot{W}}_{c,j}}.
\end{equation}
We first derive the relationship between $\vect{H}_{i,j}$ and $\dot{\vect{H}}_{i,j}$:
\begin{equation}
    \vect{H}_{i,j} = \dot{\vect{H}}_{i,j} - \dfrac{1}{I}\sum_{k=1}^I(\dot{\vect{H}}_{i,k} + \dot{\vect{H}}_{k,j}) + \dfrac{1}{I^2}\sum_{p=1}^I\sum_{q=1}^I\dot{\vect{H}}_{p,q}.
\end{equation}
Note that
\begin{equation}
    \sum_{i=1}^I\sum_{j=1}^I\vect{H}_{i,j}=0.
\end{equation}
Therefore, Eq.~\ref{eq:9} not only zero-centers the feedforwarding outputs and the back-propagated gradients, but also the Hessian matrix.
Next, we compute its Frobenius norm:
\begin{align}
\begin{split}
    \lVert \vect{H} \rVert_{F} =& \sum_{i=1}^I\sum_{j=1}^I\vect{H}^2_{i,j} \\
    =& \lVert \vect{\dot{H}} \rVert_{F} + \dfrac{1}{I^2}\big(\sum_{i=1}^I\sum_{j=1}^I\vect{\dot{H}}_{i,j}\big)^2 \\
    &- \dfrac{1}{I}\sum_{i=1}^I\Big(\sum_{j=1}^I \vect{\dot{H}}_{i,j}\Big)^2 -  \dfrac{1}{I}\sum_{j=1}^I\Big(\sum_{i=1}^I \vect{\dot{H}}_{i,j}\Big)^2 \\
    \leq & \lVert \vect{\dot{H}} \rVert_{F} - \dfrac{1}{I^2} \big(\sum_{i=1}^I\sum_{j=1}^I\vect{\dot{H}}_{i,j}\big)^2\label{eq:27}.
\end{split}
\end{align}
As shown in Eq.~\ref{eq:27}, Eq.~\ref{eq:9} reduces the Frobenius norm of the Hessian matrix by at least $\big(\sum_{i=1}^I\sum_{j=1}^I\vect{\dot{H}}_{i,j}\big)^2/I^2$, which makes the gradients more predictable than directly optimizing on the weights of the convolutional layer.

\subsection{Connections to constrained optimization}
WS imposes constraints to the weight $\hat{\vect{W}}_{c,\cdot}$ such that
\begin{equation}\label{eq:pgd_constraint}
    \sum_{i=1}^I \hat{\vect{W}}_{c,i}=0,~~~\sum_{i=1}^I \hat{\vect{W}}^2_{c,i}=I,~~~\forall c.
\end{equation}
Therefore, an alternative to the proposed WS is to consider the problem as constrained optimization and uses Projected Gradient Descent (PGD) to find the solution.
The update rule for PGD can be written as
\begin{align}\label{eq:pgd_update}
\hat{\vect{W}}^{t+1}_{c,i} = \texttt{Proj}\big(\hat{\vect{W}}^t_{c,i} - \epsilon\cdot\nabla_{\hat{\vect{W}}^t_{c,i}}\mathcal{L}\big)\\\nonumber
\end{align}
where $\texttt{Proj}(\cdot)$ denotes the projection function and $\epsilon$ denotes the learning rate.
To satisfy Eq.~\ref{eq:pgd_constraint}, $\texttt{Proj}(\cdot)$ standardizes its input. We can approximate the right hand side of Eq.~\ref{eq:pgd_update} by minimizing the \textbf{Lagrangian} of the loss function $\mathcal{L}$, which obtains
\begin{align}\label{eq:pgd_update_lagrangian}
\hat{\vect{W}}^{t+1}_{c,i}&\approx \hat{\vect{W}}^t_{c,i} - \epsilon\Big(\nabla_{\hat{\vect{W}}^t_{c,i}}\mathcal{L} - \langle \hat{\vect{W}}_{c,\cdot},\nabla_{\hat{\vect{W}}_{c,\cdot}}\mathcal{L} \rangle\hat{\vect{W}}_{c,i}\\\nonumber&-\frac{1}{I}\langle \mathbf{1},  \nabla_{\hat{\vect{W}}_{c,\cdot}}\mathcal{L}\rangle\Big)
\end{align}
Different from Eq.~\ref{eq:pgd_update}, the update rule of WS is
\begin{align}\label{eq:ws_update}
\begin{split}
    \hat{\vect{W}}^{t+1}_{c,i} =&  \texttt{Proj}\Big( \vect{W}^{t}_{c,i} - \epsilon\cdot\nabla_{{\vect{W}}^t_{c,i}}\mathcal{L}\Big) \\
    =& \texttt{Proj}\Big( \vect{W}^{t}_{c,i} - \dfrac{\epsilon}{\sigma_{\vect{W}_{c,\cdot}}} \big(  \nabla_{\hat{\vect{W}}_{c,i}}\mathcal{L} - \dfrac{1}{I} \langle \hat{\vect{W}}_{c,\cdot}, \\
    &\nabla_{\hat{\vect{W}}_{c,\cdot}}\mathcal{L} \rangle \hat{\vect{W}}_{c,i} -  \dfrac{1}{I} \langle \mathbf{1},  \nabla_{\hat{\vect{W}}_{c,\cdot}}\mathcal{L} \rangle \big) \\
    &-\dfrac{\epsilon}{I^2\cdot\sigma_{\vect{W}_{c,\cdot}}} \big\langle \mathbf{1}, \langle \hat{\vect{W}}_{c,\cdot},
    \nabla_{\hat{\vect{W}}_{c,\cdot}}\mathcal{L} \rangle \hat{\vect{W}}_{c,\cdot} \big\rangle \Big).
\end{split}
\end{align}

Eq.~\ref{eq:ws_update} is more complex than Eq.~\ref{eq:pgd_update_lagrangian}, but the increased complexity is neglectable compared with training deep networks.
For simplicity, Eq.~\ref{eq:ws_update} reuses $\texttt{Proj}$ to denote the standardization process, despite that WS uses Stochastic Gradient Descent instead of Projected Gradient Descent to optimize the weights.
\section{WS's effects on elimination singularities}

In this section, we will provide the background of BN, GN and LN, discuss the negative correlation between the performance and the distance to elimination singularities, and show LN and GN are unable to keep the networks away from elimination singularities as BN.
Next, we will show that WS helps avoiding elimination singularities.

\subsection{Batch- and channel-based normalizations and their effects on elimination singularities}

\subsubsection{Batch- and channel-based normalizations}
Based on how activations are normalized, we group the normalization methods into two types: batch-based normalization and channel-based normalization, where the batch-based normalization method corresponds to BN and the channel-based normalization methods include LN and GN.

Suppose we are going to normalize a 2D feature map $\vect{X}\in\mathds{R}^{B\times C\times H\times W}$, where $B$ is the batch size, $C$ is the number of channels, $H$ and $W$ denote the height and the width.
For each channel $c$, BN normalizes $\vect{X}$ by
\begin{equation}\label{eq:bn}
    \vect{Y}_{\cdot c \cdot\cdot} = \dfrac{\vect{X}_{\cdot c\cdot\cdot} - \mu_{\cdot c\cdot\cdot}}{\sigma_{\cdot c\cdot\cdot}},
\end{equation}
where $\mu_{\cdot c\cdot\cdot}$ and $\sigma_{\cdot c\cdot\cdot}$ denote the mean and the standard deviation of all the features of the channel $c$, $\vect{X}_{\cdot c\cdot\cdot}$.
Throughout the paper, we use $\cdot$ in the subscript to denote all the features along that dimension for convenience.

Unlike BN which computes statistics on the batch dimension in addition to the height and width, channel-based normalization methods (LN and GN) compute statistics on the channel dimension.
Specifically, they divide the channels to several groups, and normalize each group of channels, \textit{i.e.},
$\vect{X}$ is reshaped as $\vect{\dot{X}}\in\mathds{R}^{B\times G\times C/G\times H\times W}$, and then:
\begin{equation}\label{eq:cn}
    \vect{\dot{Y}}_{bg\cdot\cdot\cdot} = \dfrac{\vect{\dot{X}}_{bg\cdot\cdot\cdot} - \mu_{bg\cdot\cdot\cdot}}{\sigma{_{bg\cdot\cdot\cdot}}},
\end{equation}
for each sample $b$ of $B$ samples in a batch and each channel group $g$ out of all $G$ groups.
After Eq.~\ref{eq:cn}, the output $\vect{\dot{Y}}$ is reshaped as $\vect{\dot{X}}$ and denoted by $\vect{Y}$.

Both batch- and channel-based normalization methods have an optional affine transformation, \textit{i.e.},
\begin{equation}\label{eq:at}
    \vect{Z}_{\cdot c\cdot\cdot} = \gamma_c\vect{Y}_{\cdot c\cdot\cdot} + \beta_c.
\end{equation}

\subsubsection{BN avoids elimination singularities}
Here, we study the effect of BN on elimination singularities.
Since the normalization methods all have an optional affine transformation, we focus on the distinct part of BN, which normalizes all channels to zero mean and unit variance, \textit{i.e.},
\begin{equation}\label{eq:bnes}
    \mathbb{E}_{y\in Y_{\cdot c\cdot\cdot}}\big[y\big]=0,~~ \mathbb{E}_{y\in Y_{\cdot c\cdot\cdot}}\big[y^2\big]=1,~~\forall c.
\end{equation}
As a result, regardless of the weights and the distribution of the inputs, it guarantees that the activations of each channel are zero-centered with unit variance.
Therefore, each channel cannot be constantly deactivated because there are always some activations that are $>0$, nor underrepresented due to the channel having a very small activation scale compared with the others.

\begin{figure}
    \centering
    \includegraphics[width=\linewidth]{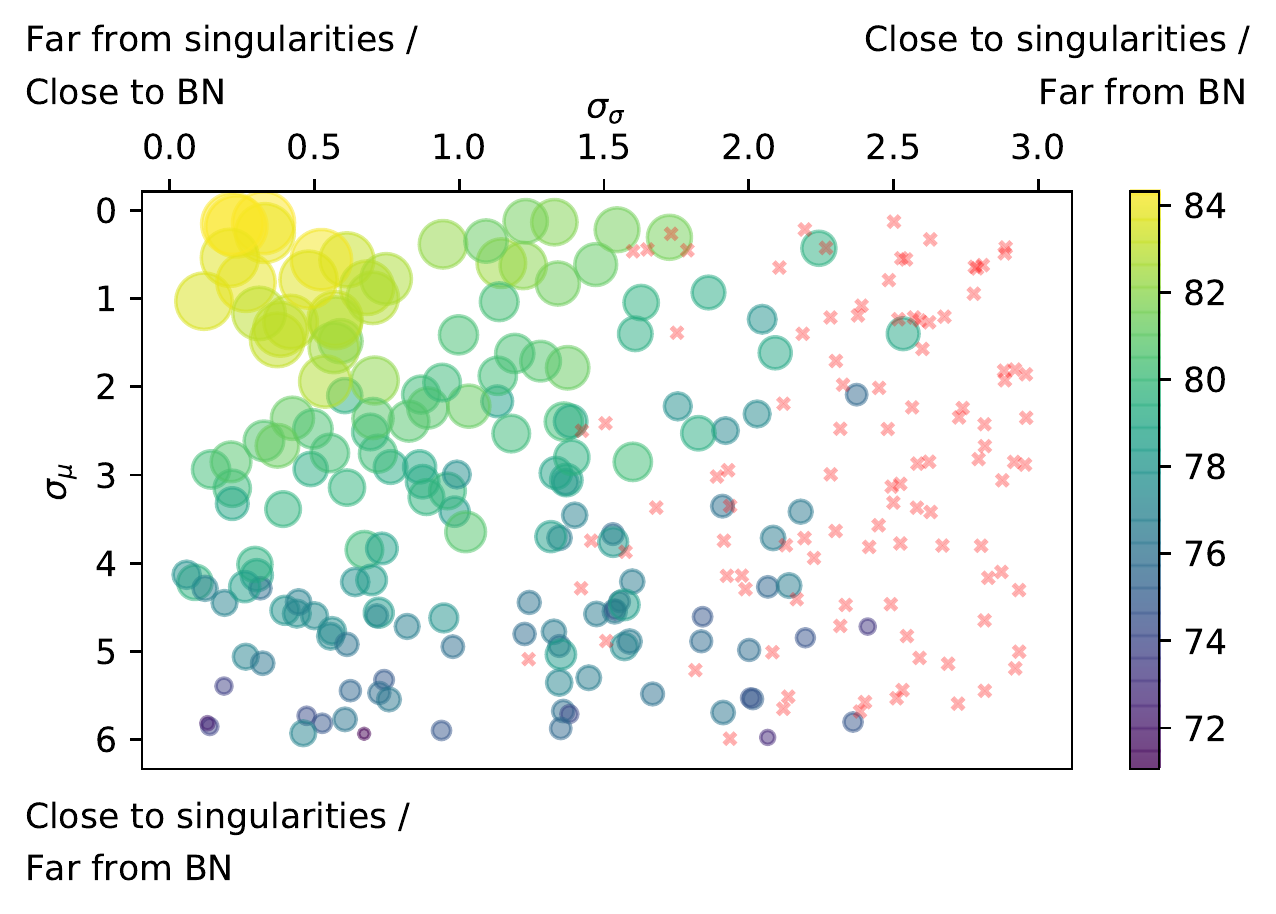}
    \caption{Model accuracy and distance to singularities.
    Larger circles correspond to higher performances.
    Red crosses represent failure cases (accuracy $<70\%$).
    Circles are farther from singularities/closer to BN if they are closer to the origin.}
    \label{fig:dist}
\end{figure}

\subsubsection{Statistical distance and its affects on performance}
\begingroup
\begin{quote}
\it
    BN avoids singularities by normalizing each channel to zero mean and unit variance.
    What if they are normalized to other means and variances?
\end{quote}
\endgroup
We ask this question because this is similar to what happens in channel-normalized models.
In the context of activation-based normalizations, BN completely resolve the issue of elimination singularities as each channel is zero-centered with unit variance.
By contrast, channel-based normalization methods, as they do \textbf{not} have batch information, are \textbf{unable} to make sure that all neurons have zero mean and unit variance after normalization.
In other words, there are likely some underrepresented channels after training if the model is using channel-based normalizations.
Since BN represents the ideal case which has the furthest distance to elimination singularities, and any dissimilarity with BN will lead to lightly or heavily underrepresented channels and thus make the models closer to singularities,
\emph{we use the distance to BN as the distance to singularities for activation-based normalizations.}
Specifically, in this definition, the model is \textit{closer} to singularities when it is \textit{far} from BN.
Fig.~\ref{fig:dist} shows that this definition is useful, where we study the relationship between the performance and the distance to singularities (\textit{i.e.}, how far from BN) caused by statistical differences.
We conduct experiments on a 4-layer convolutional network, the results of which are shown in Fig~\ref{fig:dist}.
Each convolutional layer has 32 output channels, and is followed by an average pooling layer which down-samples the features by a factor of 2.
Finally, a global average pooling layer and a fully-connected layer output the logits for Softmax.
The experiments are done on CIFAR-10~\cite{cifar}.

In the experiment, each channel $c$ will be normalized to a pre-defined mean $\hat{\mu}_{c}$ and a pre-defined variance $\hat{\sigma}_{c}$ that are drawn from two distributions, respectively:
\begin{equation}
    \hat{\mu}_{c}\sim\mathcal{N}(0, \sigma_{\mu}) ~~\text{and}~~ \hat{\sigma}_{c}=e^{\dot{\sigma}_{c}}~\text{where}~ \dot{\sigma}_{c}\sim\mathcal{N}(0,\sigma_\sigma).
\end{equation}
\textit{The model will be closer to singularities when $\sigma_{\mu}$ or $\sigma_{\sigma}$ increases.
BN corresponds to the case where $\sigma_{\mu}=\sigma_{\sigma}=0$}.

After getting $\hat{\mu}_{c}$ and $\hat{\sigma}_{c}$ for each channel, we compute
\begin{equation}
    \vect{Y}_{\cdot c\cdot\cdot} = \gamma_c\big(\hat{\sigma}_{c}\dfrac{\vect{X}_{\cdot c\cdot\cdot} - \mu_{\cdot c\cdot\cdot}}{\sigma_{\cdot c\cdot\cdot}} + \hat{\mu}_{c}\big) + \beta_c.
\end{equation}
Note that $\hat{\mu}_{c}$ and $\hat{\sigma}_{c}$ are fixed during training while $\gamma_c$ and $\beta_c$ are trainable parameters in the affine transformation.

\begin{figure}
    \centering
    \includegraphics[width=\linewidth]{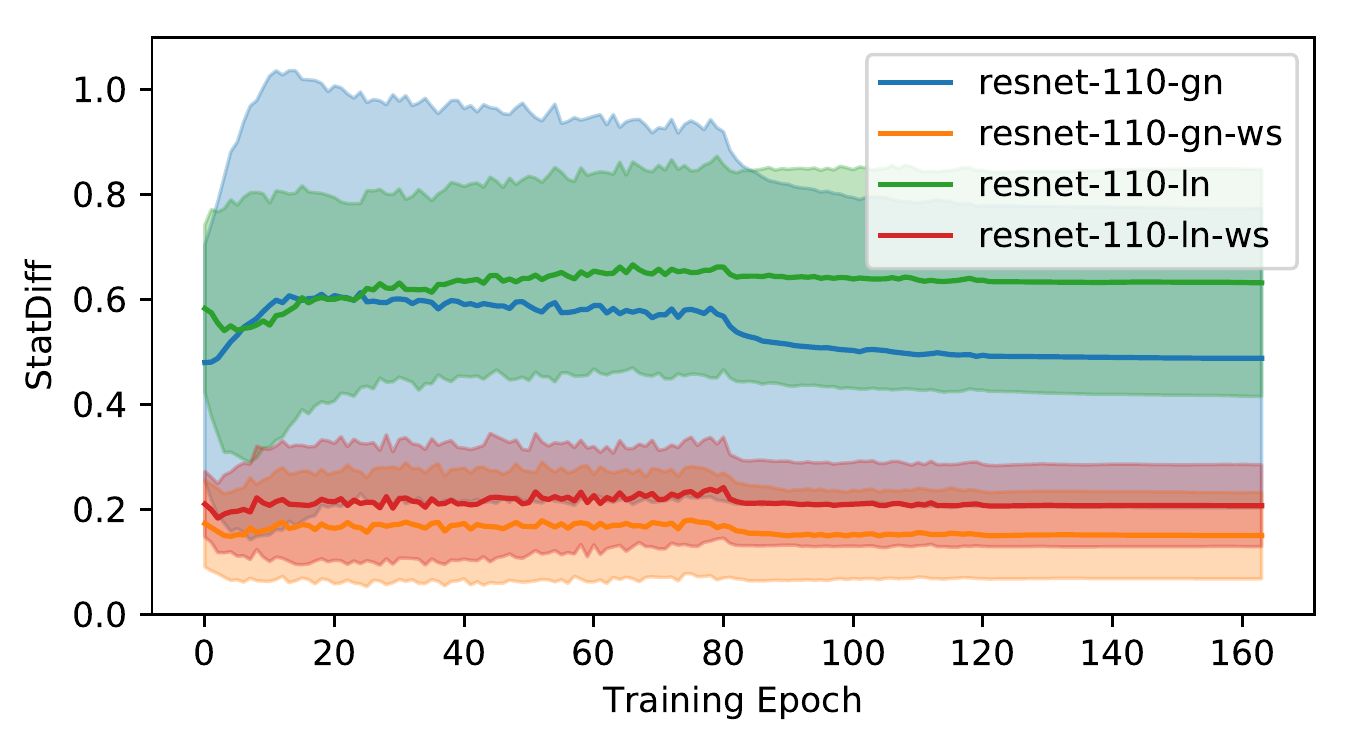}
    \caption{
    Means and standard deviations of the statistical differences (StatDiff defined in Eq.~\ref{eq:sd}) of all layers in a ResNet-110 trained on CIFAR-10 with GN, GN+WS, LN, and LN+WS.
    }
    \label{fig:stat_diff}
\end{figure}

Fig.~\ref{fig:dist} shows the experimental results.
When $\sigma_{\mu}$ and $\sigma_{\sigma}$ are closer to the origin, the normalization method is more close to BN.
When their values increase, we observe performance decreases.
For extreme cases, we also observe training failures.
These results indicate that although the affine transformation theoretically can find solutions that cancel the negative effects of normalizing channels to different statistics, their capability is limited by the gradient-based training.
They show that defining distance to singularities as the distance to BN is useful.
They also raise concerns about channel normalizations regarding their distances.

\subsubsection{Statistics in Channel-based Normalization}
Following our concerns about channel-based normalization and their distance to singularities, we study the statistical differences between channels when they are normalized by a channel-based normalization such as GN or LN.

\noindent\textbf{Statistical differences in GN, LN and WS:}
We train a ResNet-110~\cite{resnet} on CIFAR-10~\cite{cifar} normalized by GN, LN, with and without WS.
During training, we keep record of the running mean $\mu^r_c$ and variance $\sigma_c^r$ of each channel $c$ after convolutional layers.
For each group $g$ of the channels that are normalized together, we compute their channel \textbf{statistical difference} defined as the standard deviation of their means divided by the mean of their standard deviations, \textit{i.e.},
\begin{equation}\label{eq:sd}
    \text{StatDiff}(g) = \dfrac{ \sqrt{\mathbb{E}_{c\in g}\big[(\mu^r_{c})^2\big] - \big(\mathbb{E}_{c\in g}\big[\mu^r_{c}\big]\big)^2}}{\mathbb{E}_{c\in g}\big[ \sigma_{c} \big]}.
\end{equation}
We plot the average statistical differences of all the groups after every training epoch as shown in Fig.~\ref{fig:stat_diff}.

\begin{figure}
    \centering
    \begin{subfigure}[b]{\linewidth}
        \centering
         \includegraphics[width=\linewidth]{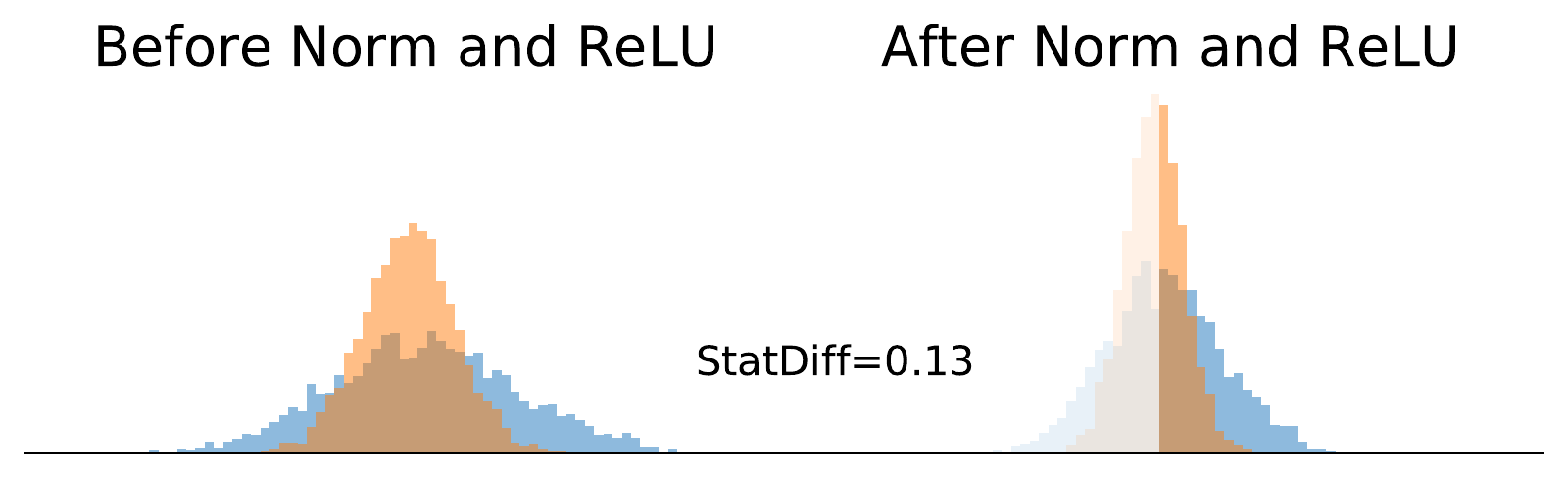}
    \end{subfigure}
    \begin{subfigure}[b]{\linewidth}
        \centering
         \includegraphics[width=\linewidth]{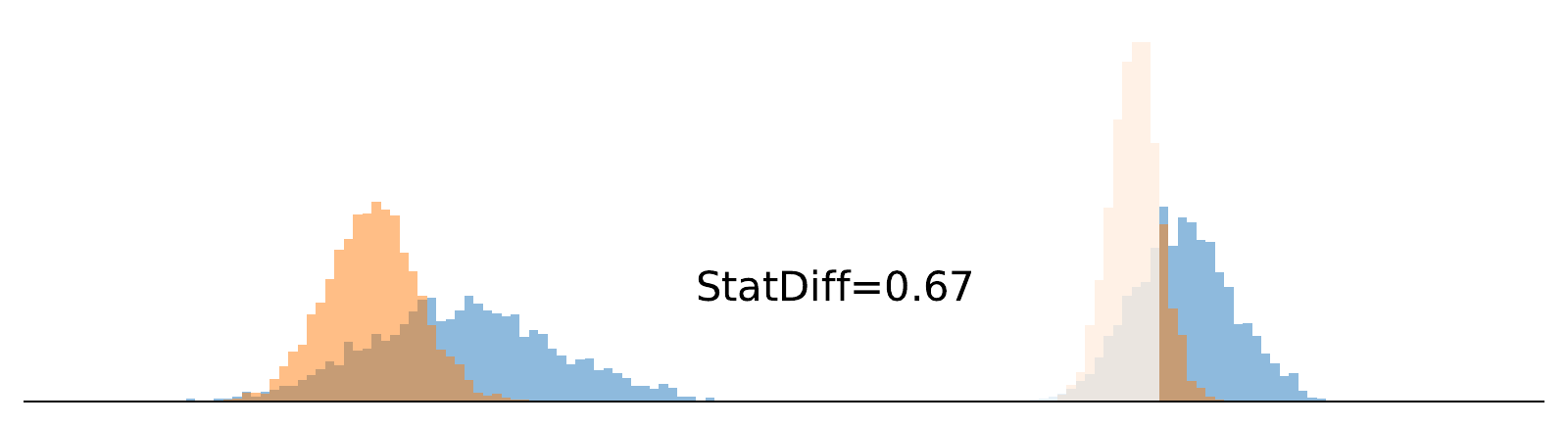}
    \end{subfigure}
    \begin{subfigure}[b]{\linewidth}
        \centering
         \includegraphics[width=\linewidth]{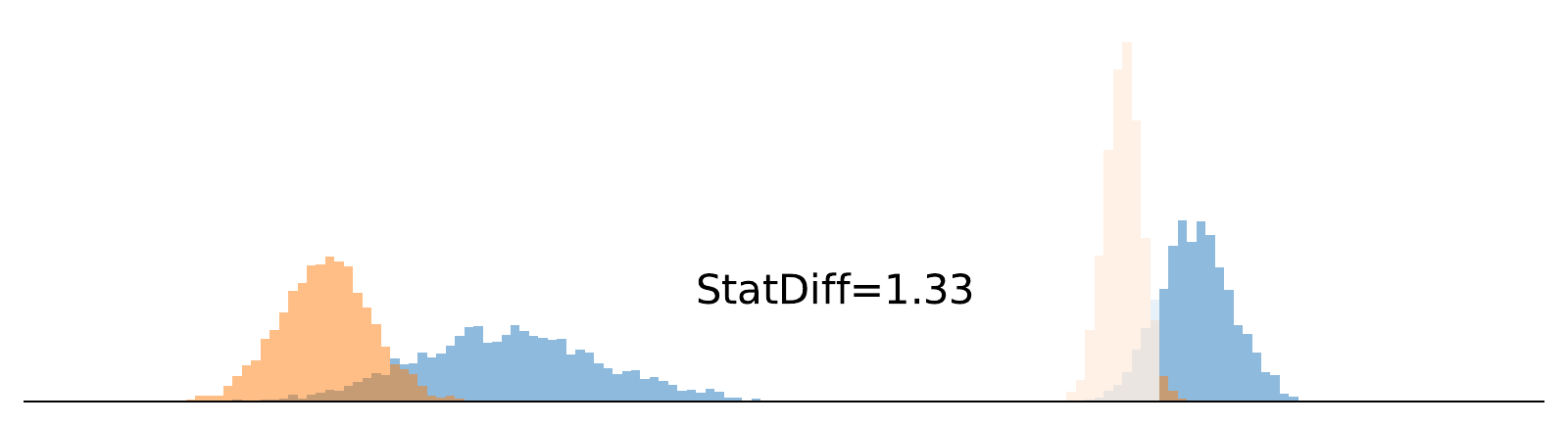}
    \end{subfigure}
    \caption{Examples of normalizing two channels in a group when they have different means and variances.
    Transparent bars mean they are $0$ after ReLU.
    StatDiff is defined in Eq.~\ref{eq:sd}.}
    \label{fig:dist_exp}
\end{figure}

By Eq.~\ref{eq:sd}, $\text{StatDiff}(g)\geq 0,~\forall g$.
In BN, all their means are the same, as well as their variances, thus $\text{StatDiff}(g)=0$.
As the value of $\text{StatDiff}(g)$ goes up, the differences between channels within a group become larger.
Since they will be normalized together as in Eq.~\ref{eq:cn}, large differences will inevitably lead to underrepresented channels.
Fig.~\ref{fig:dist_exp} plots 3 examples of 2 channels before and after normalization in Eq.~\ref{eq:cn}.
Compared with those examples, it is clear that the models in Fig.~\ref{fig:stat_diff} have many underrepresented channels.

\noindent\textbf{Why GN performs better than LN:}
Fig.~\ref{fig:stat_diff} also provides explanations why GN performs better than LN.
Comparing GN and LN, the major difference is their numbers of groups for channels: LN has only one group for all the channels in a layer while GN collects them into several groups.
A strong benefit of having more than one group is that it guarantees that each group will at least have one neuron that is not suppressed by the others from the same group.
Therefore, GN provides a mechanism to prevent the models from getting too close to singularities.

Fig.~\ref{fig:stat_diff} also shows the statistical differences when WS is used.
From the results, we can clearly see that WS makes StatDiff much closer to $0$.
Consequently, the majority of the channels are not underrepresented in WS: most of them are frequently activated and they are at similar activation scales.
This makes training with WS easier and their results better.

\subsection{WS helps avoiding elimination singularities}
The above discussions show that WS helps keeping models away from elimination singularities.
Here, we discuss why WS is able to achieve this.
Recall that WS adds constraints to the weight $\vect{W}\in\mathds{R}^{\text{O}\times\text{I}}$ of a convolutional layer with O output channels and I inputs such that $\forall c$,
\begin{equation}\label{eq:ws1}
    \sum_{i=1}^I\vect{W}_{c,i} = 0,~~~\sum_{i=1}^I\vect{W}^2_{c,i} = 1.
\end{equation}
With the constraints of WS, $\mu_{c}^{\text{out}}$ and $\sigma_{c}^{\text{out}}$ become
\begin{equation}\label{eq:ws2}
    \mu_{c}^{\text{out}}=\sum_{i=1}^I\vect{W}_{c,i}\mu_i^{\text{in}},~~~(\sigma_{c}^{\text{out}})^2=\sum_{i=1}^I\vect{W}_{c,i}^2(\sigma_i^{\text{in}})^2,
\end{equation}
when we follow the assumptions in Xavier initialization~\cite{glorot2010understanding}.
When the input channels are similar in their statistics, \textit{i.e.}, $\mu_{i}^{\text{in}}\approx\mu_{j}^{\text{in}}$, $\sigma_{i}^{\text{in}}\approx\sigma_{j}^{\text{in}}$, $\forall i,j$,
\begin{eqnarray}
    \mu_{c}^{\text{out}}&\approx&\mu_1^{\text{in}}\sum_{i=1}^I\vect{W}_{c,i}=0, \\
    (\sigma_{c}^{\text{out}})^2&\approx&(\sigma_{1}^{\text{in}})^2\sum_{i=1}^I\vect{W}_{c,i}^2=(\sigma_{1}^{\text{in}})^2.
\end{eqnarray}
In other words, WS can pass the statistical similarities from the input channels to the output channels, all the way from the image space where RGB channels are properly normalized.
This is similar to the objective of Xavier initialization~\cite{glorot2010understanding} or Kaiming initialization~\cite{he2015delving}, except that WS enforces it by reparameterization throughout the entire training process, thus is able to reduce the statistical differences a lot, as shown in Fig.~\ref{fig:stat_diff}.

Here, we summarize this subsection.
We have shown that channel-based normalization methods, as they do not have batch information, are not able to ensure a far distance from elimination singularities.
Without the help of batch information, GN alleviates this issue by assigning channels to more than one group to encourage more activated neurons, and WS adds constraints to pull the channels to be not so statistically different.
We notice that the batch information is not hard to collect in reality.
This inspires us to equip channel-based normalization with batch information, and the result is Batch-Channel Normalization.

\section{Batch-Channel Normalization}\label{sec:norm}

The previous section discusses elimination singularities and shows WS is able to keep models away from them.
To fully address the issue of elimination singularities, we propose Batch-Channel Normalization (BCN).
This section presents the definition of BCN, discusses why adding batch statistics to channel normalization is not redundant, and shows how BCN runs in large-batch and micro-batch training settings.

\subsection{Definition}
Batch-Channel Normalization (BCN) adds batch information and constraints to channel-based normalization methods.
Let $\vect{X}\in\mathds{R}^{B\times C\times H\times W}$ be the features to be normalized.
Then, the normalization is done as follows.
$\forall c$,
\begin{equation}\label{eq:bcnbn}
    \vect{\dot{X}}_{\cdot c\cdot\cdot}=\gamma_c^b \dfrac{\vect{X}_{\cdot c\cdot\cdot} - \hat{\mu}_{ c}}{\hat{\sigma}_{c}}+\beta_{c}^b,
\end{equation}
where the purpose of $\hat{\mu}_{c}$ and $\hat{\sigma}_{c}$ is to make
\begin{equation}\label{eq:bcn_purpose}
    \mathbb{E}\Big\{\dfrac{\vect{X}_{\cdot c\cdot\cdot} - \hat{\mu}_{ c}}{\hat{\sigma}_{c}}\Big\}= 0~\text{and}~\mathbb{E}\Big\{\big(\dfrac{\vect{X}_{\cdot c\cdot\cdot} - \hat{\mu}_{ c}}{\hat{\sigma}_{c}}\big)^2\Big\}= 1.
\end{equation}
Then, $\vect{\dot{X}}$ is reshaped as $\vect{\dot{X}}\in\mathds{R}^{B\times G\times C/G\times H\times W}$ to have $G$ groups of channels.
Next, $\forall g, b$,
\begin{equation}\label{eq:bcngn}
    \vect{\dot{Y}}_{bg\cdot\cdot\cdot} = \gamma_g^c\dfrac{\vect{\dot{X}}_{bg\cdot\cdot\cdot} - \mu_{bg\cdot\cdot\cdot}}{\sigma{_{bg\cdot\cdot\cdot}}}+ \beta_g^c.
\end{equation}
Finally, $\vect{\dot{Y}}$ is reshaped back to $\vect{Y}\in\mathds{R}^{B\times C\times H\times W}$, which is the output of the Batch-Channel Normalization.

\subsection{Large- and Micro-batch Implementations}
Note that in Eq.~\ref{eq:bcnbn} and \ref{eq:bcngn}, only two statistics need batch information: $\hat{\mu}_{c}$ and $\hat{\sigma}_{c}$, as their values depend on more than one sample.
Depending on how we obtain the values of $\hat{\mu}_{c}$ and $\hat{\sigma}_{c}$, we have different implementations for large-batch and micro-batch training settings.

\subsubsection{Large-batch training}
When the batch size is large, estimating $\hat{\mu}_{c}$ and $\hat{\sigma}_{c}$ is easy: we just use a Batch Normalization layer to achieve the function of Eq.~\ref{eq:bcnbn} and \ref{eq:bcn_purpose}.
As a result, the proposed BCN can be written as
\begin{equation}
    \text{BCN}(\vect{X})=\text{CN}(\text{BN}(\vect{X})).
\end{equation}
Implementing it is also easy with modern deep learning libraries, which is omitted here.

\subsubsection{Micro-batch training}
One of the motivations of channel normalization is to allow deep networks to train on tasks where the batch size is limited by the GPU memory.
Therefore, it is important for Batch-Channel Normalization to be able to work in the micro-batch training setting.

\begin{algorithm}[t]
\SetAlgoLined
\SetKwInput{KwInput}{Input}
\SetKwInput{KwOutput}{Output}
 \KwInput{$\vect{X}\in\mathds{R}^{B\times C\times H\times W}$, the current estimates of $\hat{\mu}_c$ and $\hat{\sigma}^2_c$, and the update rate $r$.}
 \KwOutput{Normalized $\vect{Y}$.}
 Compute $\dot{\mu}_{c}\leftarrow\frac{1}{BHW}\sum_{b,h,w}\vect{X}_{b,c,h,w}$\;
 Compute $\dot{\sigma}^2_{c}\leftarrow\frac{1}{BHW}\sum_{b,h,w} \big( \vect{X}_{b,c,h,w} - \hat{\mu}_c \big)^2$\;
 Update $\hat{\mu}_c\leftarrow \hat{\mu}_c + r (\dot{\mu}_c - \hat{\mu}_c)$\;
 Update $\hat{\sigma}^2_c\leftarrow \hat{\sigma}^2_c + r (\dot{\sigma}^2_c - \hat{\sigma}^2_c)$\;
 Normalize $\vect{\dot{X}}_{\cdot c\cdot\cdot}=\gamma_c^b \dfrac{\vect{X}_{\cdot c\cdot\cdot} - \hat{\mu}_{ c}}{\hat{\sigma}_{c}}+\beta_{c}^b$\;
 Reshape $\vect{\dot{X}}$ to $\vect{\dot{X}}\in\mathds{R}^{B\times G\times C/G\times H\times W}$\;
 Normalize $\vect{\dot{Y}}_{bg\cdot\cdot\cdot} = \gamma_g^c\dfrac{\vect{\dot{X}}_{bg\cdot\cdot\cdot} - \mu_{bg\cdot\cdot\cdot}}{\sigma{_{bg\cdot\cdot\cdot}}}+ \beta_g^c$\;
 Reshape $\vect{\dot{Y}}$ to $\vect{{Y}}\in\mathds{R}^{B\times C\times H\times W}$\;
 \caption{Micro-batch BCN}\label{alg:1}
\end{algorithm}

Algorithm~\ref{alg:1} shows the feed-forwarding implementation of the micro-batch Batch-Channel Normalization.
The basic idea behind this algorithm is to constantly estimate the values of $\hat{\mu}_c$ and $\hat{\sigma}_c$, which are initialized as $0$ and $1$, respectively, and normalize $\vect{X}$ based on these estimates.
It is worth noting that in the algorithm, $\hat{\mu}_c$ and $\hat{\sigma}_c$ are not updated by the gradients computed from the loss function;
instead, they are updated towards more accurate estimates of those statistics.
Step 3 and 4 in Algorithm~\ref{alg:1} resemble the update steps in gradient descent; thus, the implementation can also be written in gradient descent by storing the difference $\Delta\hat{\mu}_c$ and $\Delta\hat{\sigma}_c$ as their gradients.
Moreover, we set the update rate $r$ to be the learning rate of trainable parameters.

Algorithm~\ref{alg:1} also raises an interesting question: when researchers study the micro-batch issue of BN before, why not just using the estimates to batch-normalize the features?
In fact, \cite{batchrenorm} tries a similar idea, but does not fully solve the micro-batch issue: it needs a bootstrap phase to make the estimates meaningful, and the performances are usually not satisfactory.
The underlying difference between micro-batch BCN and \cite{batchrenorm} is that BCN has a channel normalization following the estimate-based normalization.
This makes the previously unstable estimate-based normalization stable, and the reduction of Lipschitz constants which speeds up training is also done in the channel-based normalization part, which is also impossible to do in estimate-based normalization.
In summary, \textit{channel-based normalization makes estimate-based normalization possible, and estimate-based normalization helps channel-based normalization to keep models away from elimination singularities}. 

\begin{table}[t]
\small
\setlength{\tabcolsep}{1.2em}
    \centering
    \begin{tabular}{l|c|l|c}
    \toprule
        Method & Top-1 & Method & Top-1 \\
    \midrule
        LN~\cite{layernorm} & 27.22 & LN+WS & 24.60 \\
        IN~\cite{instnorm} & 29.49 & IN+WS & 28.24 \\
        GN~\cite{groupnorm} & 24.81 & GN+WS & 23.72 \\
        BN~\cite{batchnorm} & 24.30 & BN+WS & 23.76 \\
    \bottomrule
    \end{tabular}
    \vspace{0.05in}
    \caption{Top-1 error rates of ResNet-50 on ImageNet.
    All models except BN are trained with batch size 1 per GPU.
    BN models are trained with batch size 64 per GPU.}
    \label{tab:abl}
\end{table}

\subsection{Is Batch-Channel Normalization Redundant?}
Batch- and channel-based normalizations are similar in many ways.
Is BCN thus redundant as it normalizes normalized features?
Our answer is \textbf{no}.
Channel normalizations need batch knowledge to keep the models away from elimination singularities; at the same time, it also brings benefits to the batch-based normalization, including:

{\noindent\bf Batch knowledge without large batches.}
Since BCN runs in both large-batch and micro-batch settings, it provides a way to utilize batch knowledge to normalize activations without relying on large training batch sizes.

{\noindent\bf Additional non-linearity.}
Batch Normalization is linear in the test mode or when the batch size is large in training.
By contrast, channel-based normalization methods, as they normalize each sample individually, are not linear.
They will add strong non-linearity and increase the model capacity.

{\noindent\bf Test-time normalization.}
Unlike BN that relies on estimated statistics on the training dataset for testing, channel normalization normalizes testing data again, thus allows the statistics to adapt to different samples.
As a result, channel normalization will be more robust to statistical changes and show better generalizability for unseen data. 

\section{Experimental Results}\label{sec:exp}

\begin{table*}[t]
\small
    \centering
    \begin{tabular}{l|cc|cc|cc|cc|cc}
    \toprule
        Method -- Batch Size & \multicolumn{2}{c|}{BN~\cite{batchnorm} -- 64 / 32} & \multicolumn{2}{c|}{SN~\cite{switchnorm} -- 1} &  \multicolumn{2}{c|}{GN~\cite{groupnorm} -- 1} & 
        \multicolumn{2}{c|}{BN+WS -- 64 / 32} &
        \multicolumn{2}{c}{GN+WS -- 1} \\
        \cmidrule{2-11}
        & Top-1 & Top-5 & Top-1 & Top-5 & Top-1 & Top-5 & Top-1 & Top-5 & Top-1 & Top-5 \\
        \midrule
        ResNet-50~\cite{resnet} & 24.30 & 7.19 & 25.00 & -- & 24.81 & 7.46 & 23.76 & 7.13 & 23.72 & 6.99 \\
        ResNet-101~\cite{resnet} & 22.44 & 6.21 & -- & -- & 22.87 & 6.51 & 21.89 & 6.01 & 22.10 & 6.07 \\
        \bottomrule
    \end{tabular}
    \vspace{0.05in}
    \caption{Error rates of ResNet-50 and ResNet-101 on ImageNet.
    ResNet-50 models with BN are trained with batch size 64 per GPU, and ResNet-101 models with BN are trained with 32 images per GPU.
    The others are trained with 1 image per GPU. }
    \label{tab:imagenet_ws}
\end{table*}

In this section, we will present the experimental results of using our proposed Weight Standardization and Batch-Channel Normalization, including image classification on CIFAR-10/100~\cite{cifar} and ImageNet~\cite{ILSVRC15}, object detection and instance segmentation on COCO~\cite{coco}, video
recognition on Something-SomethingV1 dataset~\cite{something}, and semantic segmentation on PASCAL VOC~\cite{pascal}.

\subsection{Image Classification on ImageNet}

\subsubsection{Weight Standardization}
ImageNet is a large-scale image classification dataset.
There are about 1.28 million training samples and 50K validation images.
It has 1000 categories, each has roughly 1300 training images and exactly 50 validation samples.

Table~\ref{tab:abl} shows the top-1 error rates of ResNet-50 on ImageNet when it is trained with different normalization methods, including Layer Normalization~\cite{layernorm}, Instance Normalization~\cite{instnorm}, Group Normalization~\cite{groupnorm} and Batch Normalization.
From Table~\ref{tab:abl}, we can see that when the batch size is limited to 1, GN+WS is able to achieve performances comparable to BN with large batch size.
Therefore, we will use GN+WS for micro-batch training because GN shows the best results among all the normalization methods that can be trained with 1 image per GPU.

Table~\ref{tab:imagenet_ws} shows our major experimental results of WS on the ImageNet dataset~\cite{ILSVRC15}.
Note that Table~\ref{tab:imagenet_ws} only shows the error rates of ResNet-50 and ResNet-101.
This is to compare with the previous work that focus on micro-batch training problem, e.g. Switchable Normalization~\cite{switchnorm} and Group Normalization~\cite{groupnorm}.
We run all the experiments using the official PyTorch implementations of the layers except for SN~\cite{switchnorm} which are the performances reported in their paper.
This makes sure that all the experimental results are comparable, and our improvements are reproducible.

\begin{table}[t]
\small
\setlength{\tabcolsep}{0.8em}
    \centering
    \begin{tabular}{l|ccc|c}
    \toprule
        Backbone & WN & CWN & WS & Top-1 \\
    \midrule
        ResNet-50 + GN & & & & 24.81 \\
        ResNet-50 + GN & \cmark & & & 25.09 \\
        ResNet-50 + GN & & \cmark & & 24.23 \\
        ResNet-50 + GN & & & \cmark & 23.72 \\
    \bottomrule
    \end{tabular}
    \vspace{0.05in}
    \caption{Comparing Top-1 error rates between WS, WN and CWN on ImageNet.
    The backbone is a ResNet-50 normalized by GN and trained with batch size 1 per GPU.}
    \label{tab:cwn}
\end{table}

Table~\ref{tab:cwn} compares WS with other weight-based normalization methods including WN and CWN.
To show the comparisons, we train the same ResNet-50 normalized by GN on activations with different weight-based normalizations.
The code of WN uses the official PyTorch implementation, and the code of CWN is from the official implementation of their github.
From the results, we can observe that these normalization methods have different effects on the performances of the models.
Compared with WN and CWN, the proposed WS achieves lower top-1 error rate.

Table~\ref{tab:ind_abl} shows the individual effects of Eq.~\ref{eq:9} and \ref{eq:10} on training deep neural networks.
Consistent with Fig.~\ref{fig:mean_div}, Eq.~\ref{eq:9} is the major component that brings performance improvements.
These results are also consistent with the theoretical results we have on the Lipschitz analysis.

\begin{table}[t]
\small
\setlength{\tabcolsep}{0.8em}
    \centering
    \begin{tabular}{l|cc|c}
    \toprule
        Backbone & - mean & / std & Top-1 \\
    \midrule
        ResNet-50 + GN & & & 24.81 \\
        ResNet-50 + GN & \cmark & & 23.96  \\
        ResNet-50 + GN & & \cmark & 24.60 \\
        ResNet-50 + GN & \cmark & \cmark & 23.72 \\
    \bottomrule
    \end{tabular}
    \vspace{0.05in}
    \caption{Comparing Top-1 error rates between WS (``- mean": Eq.~\ref{eq:9}, and ``/ div": Eq.~\ref{eq:10}) and its individual effects.
    The backbone is a ResNet-50-GN trained with batch size 1 per GPU.}
    \label{tab:ind_abl}
\end{table}

\begin{table}[t]
\small
    \centering
    \begin{tabular}{l|cc|cc}
    \toprule
        Method & \multicolumn{2}{c|}{GN~\cite{groupnorm}} &
        \multicolumn{2}{c}{GN+WS~\cite{groupnorm}} \\
        \cmidrule{2-5}
        Batch Size = 1 & Top-1 & Top-5 & Top-1 & Top-5 \\
        \midrule
        ResNeXt-50~\cite{resnext} & 24.24 & 7.27 & 22.71 & 6.38\\
        ResNeXt-101~\cite{resnext} & 22.86 & 6.51 & 21.80 & 6.03\\
        \bottomrule
    \end{tabular}
    \vspace{0.05in}
    \caption{ResNeXt-50 and ResNeXt-101 on ImageNet. All models are trained with batch size 1 per GPU.}
    \label{tab:resnext}
\end{table}

In Table~\ref{tab:resnext}, we also provide the experimental results on ResNeXt~\cite{resnext}.
Here, we show the performance comparisons between ResNeXt+GN and ResNeXt+GN+WS.
Note that GN originally did not provide results on ResNeXt.
Without tuning the hyper-parameters in the Group Normalization layers, we use 32 groups for each of them which is the default configuration for ResNet that GN was originally proposed for.
ResNeXt-50 and 101 are 32x4d.
We train the models for 100 epochs with batch size set to 1 and iteration size set to 32.
As the table shows, the performance of GN on training ResNeXt is unsatisfactory: they perform closely to the original ResNets.
In the same setting, WS is able to make training ResNeXt a lot easier.

\subsubsection{Batch-Channel Normalization}

\begin{figure}
    \centering
    \includegraphics[width=\linewidth]{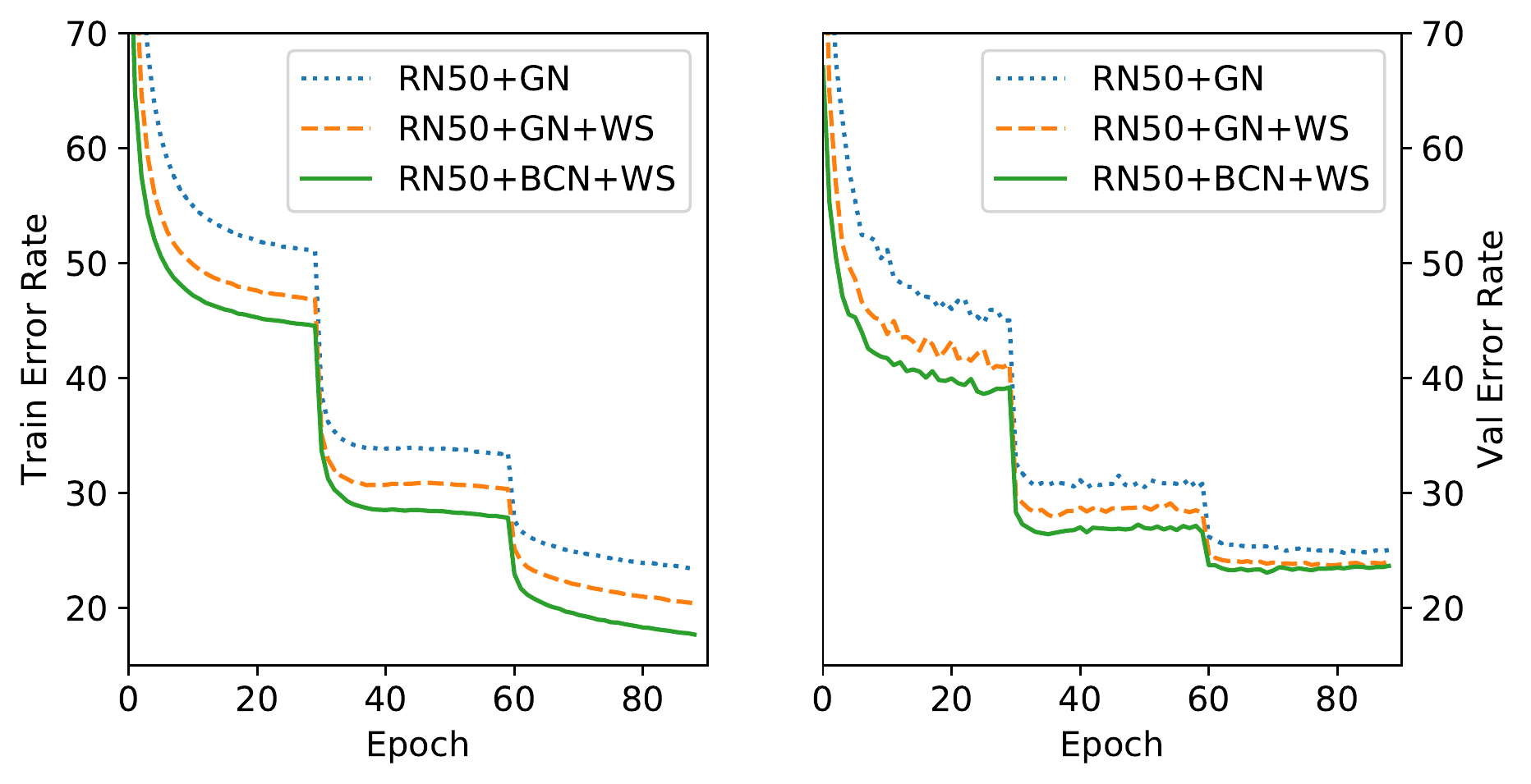}
    \caption{Training and validation error rates of ResNet-50 on ImageNet. The comparison is between the baselines GN ~\cite{groupnorm}, GN + WS, and Batch-Channel Normalization (BCN) with WS. Our method BCN and WS not only significantly improve the training speed, they also lower the error rates of the final models by a comfortable margin.}
    \label{fig:imagenet}
\end{figure}

Fig.~\ref{fig:imagenet} shows the training dynamics of ResNet-50 with GN, GN+WS and BCN+WS, and Table~\ref{tab:imagenet} shows the top-1 and top-5 error rates of ResNet-50 and ResNet-101 trained with different normalization methods.
From the results, we observe that adding batch information to channel-based normalizations strongly improves their accuracy.
As a result, GN, whose performances are similar to BN when used with WS, now is able to achieve better results than the BN baselines.
And we find improvements not only in the final model accuracy, but also in the training speed.
As shown in Fig.~\ref{fig:imagenet}, we see a big drop of training error rates at each epoch.
This demonstrates that the model is now farther from elimination singularities, resulting in an easier and faster learning.

\begin{table}[]
\small
    \centering
    \begin{tabular}{l|ccc|cc}
        \toprule
        Backbone & GN & WS & BCN & Top-1 & Top-5 \\
        \midrule
        ResNet-50 & \cmark & & & 24.81 & 7.46 \\
        ResNet-50 & \cmark & \cmark & & 23.72 & 6.99 \\
        ResNet-50 & & \cmark & \cmark & \bf23.09 & \bf6.55 \\\midrule
        ResNet-101 & \cmark & & & 22.87 & 6.51 \\
        ResNet-101 & \cmark & \cmark & & 22.10 & 6.07 \\
        ResNet-101 & & \cmark & \cmark & \bf 21.29 & \bf 5.60 \\
        \bottomrule
    \end{tabular}
    \caption{Top-1/5 error rates of ResNet-50, ResNet-101, and ResNeXt-50 on ImageNet. The test size is $224\times224$ with center cropping. All normalizations are trained with batch size $32$ or $64$ per GPU without synchronization.}
    \label{tab:imagenet}
\end{table}

\subsubsection{Experiment settings}
Here, we list the hyper-parameters used for getting all those results.
For all models, the learning rate is set to 0.1 initially, and is multiplied by $0.1$ after every 30 epochs.
We use SGD to train the models, where the weight decay is set to 0.0001 and the momentum is set to 0.9.
For ResNet-50 with BN or BN+WS, the training batch is set to 256 for 4 GPUs.
Without synchronized BN~\cite{syncbn}, the effective batch size is 64.
For other ResNet-50 where batch size is $1$ per GPU, we set the iteration size to $64$, \textit{i.e.}, the gradients are averaged across every 64 iterations and then one step is taken.
This is to ensure fair comparisons because by doing so the total numbers of parameter updates are the same even if their batch sizes are different.
We train ResNet-50 with different normalization techniques for 90 epochs.
For ResNet-101, we set the batch size to 128 because some of the models will use more than 12GB per GPU when setting their batch size to 256.
In total, we train all ResNet-101 models for 100 epochs.
Similarly, we set the iteration size for models trained with 1 image per GPU to be 32 in order to compensate for the total numbers of parameter updates.

\subsection{Image Classification on CIFAR}
CIFAR has two image datasets, CIFAR-10 (C10) and CIFAR-100 (C100).
Both C10 and C100 have color images of size $32\times 32$.
C10 dataset has 10 categories while C100 dataset has 100 categories.
Each of C10 and C100 has 50,000 images for training and 10,000 images for testing and the categories are balanced in terms of the number of samples.
In all the experiments shown here, the standard data augmentation schemes are used, \textit{i.e.}, mirroring and shifting, for these two datasets.
We also standardizes each channel of the datasets for data pre-processing.

Table~\ref{tab:cifar1} shows the experimental results that compare our proposed BCN with BN and GN.
The results are grouped into 4 parts based on whether the training is large-batch or micro-batch, and whether the dataset is C10 and C100.
On C10, our proposed BCN is better than BN on large-batch training, and is better than GN (with or without WS) which is specifically designed for micro-batch training.
Here, micro-batch training assumes the batch size is 1, and RN110 is the 110-layer ResNet~\cite{resnet} with basic block as the building block.
The number of groups here for GN is $\min\{32, (\text{the number of channels}) / 4\}$.

Table~\ref{tab:cifar2} shows comparisons with more recent normalization methods, Switchable Normalization (SN)~\cite{switchnorm} and Dynamic Normalization (DN)~\cite{dynamicnorm} which were tested for a variant of ResNet for CIFAR: ResNet-18.
To provide readers with direct comparisons, we also evaluate BCN on ResNet-18 with the group number set to $32$ for models that use GN.
Again, all the results are organized based on whether they are trained in the micro-batch setting.
Based on the results shown in Table~\ref{tab:cifar1} and \ref{tab:cifar2}, it is clear that BCN is able to outperform the baselines effortlessly in both large-batch and micro-batch training settings.

\begin{table}[]
\small
    \centering
    \begin{tabular}{ll|c|cccc|c}
        \toprule
         & Model  & Micro & BN & GN & BCN & WS & Error \\\midrule
        C10 & RN110 & & \cmark & & & & 6.43 \\
        C10 & RN110 & & & & \cmark & \cmark & 5.90 \\\midrule
        C10 & RN110 & \cmark & & \cmark & & & 7.45 \\
        C10 & RN110 & \cmark & &  \cmark & & \cmark & 6.82 \\
        C10 & RN110 & \cmark & & & \cmark & \cmark & 6.31 \\
        \midrule
        C100 & RN110 & & \cmark & & & & 28.86 \\
        C100 & RN110 & & & & \cmark & \cmark & 28.36 \\\midrule
        C100 & RN110 & \cmark & & \cmark & & & 32.86 \\
        C100 & RN110 & \cmark & &  \cmark & & \cmark & 29.49 \\
        C100 & RN110 & \cmark & & & \cmark & \cmark & 28.28 \\
        \bottomrule
    \end{tabular}
    \caption{Error rates of a 110-layer ResNet~\cite{resnet} on CIFAR-10/100~\cite{cifar} trained with BN~\cite{batchnorm}, GN~\cite{groupnorm}, and our BCN and WS.
    The results are grouped based on dataset and large/micro-batch training.
    Micro-batch assumes $1$ sample per batch while large-batch uses 128 samples in each batch.
    WS indicates whether WS is used for weights.}
    \label{tab:cifar1}
\end{table}

\begin{table}[]
\small
\setlength{\tabcolsep}{1.2em}
    \centering
    \begin{tabular}{cc|cc|c}
        \toprule
        Dataset & Model & Micro & Method & Error \\\midrule
        C10 & RN18 & & BN & 5.20 \\
        C10 & RN18 & & SN & 5.60 \\
        C10 & RN18 & & DN & 5.02 \\
        C10 & RN18 & & BCN+WS & 4.96 \\\midrule
        C10 & RN18 & \cmark & BN & 8.45 \\
        C10 & RN18 & \cmark & SN & 7.62 \\
        C10 & RN18 & \cmark & DN & 7.55 \\
        C10 & RN18 & \cmark & BCN+WS & 5.43 \\
        \hline
    \end{tabular}
    \caption{Error rates of ResNet-18 on CIFAR-10 trained with SN~\cite{switchnorm}, DN~\cite{dynamicnorm}, and our BCN and WS.
    The results are grouped based on large/micro-batch training.
    The performances of BN, SN and DN are from \cite{dynamicnorm}.
    Micro-batch for BN, SN and DN uses 2 images per batch, while BCN uses 1.}
    \label{tab:cifar2}
\end{table}

\subsection{Object Detection and Instance Segmentation}

\begin{table*}[]
\small
    \centering
    \begin{tabular}{c|ccc|ccc|ccc|ccc|ccc}
        \toprule
        Model & GN & WS & BCN & AP$^b$ & AP$^b_{.5}$ & AP$^b_{.75}$ & AP$^b_{l}$ & AP$^b_{m}$ & AP$^b_{s}$ & AP$^m$ & AP$^m_{.5}$ & AP$^m_{.75}$ & AP$^m_{l}$ & AP$^m_{m}$ & AP$^m_{s}$\\\midrule
        RN50 & \cmark & & & 39.8 & 60.5 & 43.4 & 52.4 & 42.9 & 23.0 & 36.1 & 57.4 & 38.7 & 53.6 & 38.6 & 16.9 \\
        RN50 & \cmark & \cmark & & 40.8 & 61.6 & 44.8 & 52.7 & 44.0 & 23.5 & 36.5 & 58.5 & 38.9 & 53.5 & 39.3 & 16.6\\
        RN50 & & \cmark & \cmark & \bf 41.4 & \bf 62.2 & \bf 45.2 & \bf 54.7 & \bf 45.0 & \bf 24.2 & \bf 37.3 & \bf 59.4 & \bf 39.8 & \bf 55.0 & \bf 40.1 & \bf 17.9 \\\midrule
        RN101 & \cmark & & & 41.5 & 62.0 & 45.5 & 54.8 &45.0 &24.1 & 37.0 &59.0 &39.6 & 54.5 &40.0 &17.5\\
        RN101 & \cmark & \cmark & & 42.7 & 63.6 & 46.8 & 56.0 & 46.0 & 25.7 & 37.9 & 60.4 & 40.7 & 56.3 & 40.6 & 18.2 \\
        RN101 & & \cmark & \cmark & \bf43.6 & \bf64.4 & \bf47.9 & \bf57.4 & \bf47.5 & \bf25.6 & \bf 39.1 &\bf 61.4 &\bf 42.2 &\bf 57.3 &\bf 42.1 &\bf 19.1\\
        \bottomrule
    \end{tabular}
    \caption{Object detection and instance segmentation results on COCO val2017~\cite{coco} of Mask R-CNN~\cite{maskrcnn} and FPN~\cite{fpn} with ResNet-50 and ResNet-101~\cite{resnet} as backbone.
    The models are trained with different normalization methods, which are used in their backbones, bounding box heads, and mask heads.}
    \label{tab:mask}
\end{table*}

Unlike image classification on ImageNet where we could afford large batch training when the models are not too big, object detection and segmentation on COCO~\cite{coco} usually use 1 or 2 images per GPU for training due to the high resolution.
Given the good performances of our method on ImageNet which are comparable to the large-batch BN training, we expect that our method is able to significantly improve the performances on COCO because of the training setting.

We use a PyTorch-based Mask R-CNN framework\footnote{https://github.com/facebookresearch/maskrcnn-benchmark} for all the experiments.
We take the models pre-trained on ImageNet, fine-tune them on COCO train2017 set, and test them on COCO val2017 set.
To maximize the comparison fairness, we use the models we pre-trained on ImageNet instead of downloading the pre-trained models available online.
We use 4 GPUs to train the models and apply the learning rate schedules for all models following the practice used in the Mask R-CNN framework our work is based on.
We use 1X learning rate schedule for Faster R-CNN and 2X learning rate schedule for Mask R-CNN.
For ResNet-50, we use 2 images per GPU to train the models, and for ResNet-101, we use 1 image per GPU because the models cannot fit in 12GB GPU memory.
We then adapt the learning rates and the training steps accordingly.
The configurations we run use FPN~\cite{fpn} and a 4conv1fc bounding box head.
All the training procedures strictly follow their original settings.

Table~\ref{tab:mask} reports the Average Precision for bounding box (AP$^b$) and instance segmentation (AP$^m$) and Table~\ref{tab:fast} reports the Average Precision (AP) of Faster R-CNN trained with different methods.
From the two tables, we can observe results similar to those on ImageNet.
GN has limited performance improvements when it is used on more complicated architectures such as ResNet-101 and ResNet-101.
But when we add WS to GN or use BCN, we are able to train the models much better.
The improvements become more significant when the network complexity increases.
Considering nowadays deep networks are becoming deeper and wider, having a normalization technique such as our WS will ease the training a lot without worrying about the memory and batch size issues.

\begin{table}[]
\small
\setlength{\tabcolsep}{0.35em}
    \centering
    \begin{tabular}{c|ccc|ccc|ccc}
        \toprule
        Model & GN & WS & BCN & AP$^b$ & AP$^b_{.5}$ & AP$^b_{.75}$ & AP$^b_{l}$ & AP$^b_{m}$ & AP$^b_{s}$ \\\midrule
        RN50 & \cmark & & & 38.0 & 59.1 & 41.2 & 49.5 &40.9 &22.4 \\
        RN50 & \cmark & \cmark & & 38.9 & 60.4 & 42.1 & 50.4 &42.4 &23.5 \\
        RN50 & & \cmark & \cmark & \bf 39.7 & \bf 60.9 & \bf 43.1 & \bf 51.7 & \bf 43.2 & \bf 24.0 \\\midrule
        RN101 &\cmark & & & 39.7 & 60.9 & 43.3 & 51.9 & 43.3 &23.1  \\
        RN101 & \cmark & \cmark & & 41.3 & 62.8 & 45.1 & 53.9 & 45.2 & 24.7 \\
        RN101 & & \cmark & \cmark & \bf41.8 & \bf 63.4 & \bf 45.8 & \bf 54.1 & \bf 45.6 & \bf 25.6 \\
        \bottomrule
    \end{tabular}
    \caption{Object detection results on COCO using Faster R-CNN~\cite{fasterrcnn} and
FPN with different normalization methods.}
    \label{tab:fast}
\end{table}

\subsection{Semantic Segmentation on PASCAL VOC}

\begin{table}[]
\small
    \centering
    \begin{tabular}{cc|cccc|c}
        \toprule
        Dataset & Model & GN & BN & WS & BCN & mIoU \\\midrule
        VOC Val & RN101 & \cmark & & & & 74.90 \\
        VOC Val & RN101 & \cmark & & \cmark & & 77.20 \\\midrule
        VOC Val & RN101 & & \cmark & & & 76.49 \\
        VOC Val & RN101 & & \cmark & \cmark & & 77.15 \\\midrule
        VOC Val & RN101 & & & \cmark & \cmark & 78.22 \\
        \bottomrule
    \end{tabular}
    \caption{Comparisons of semantic segmentation performance of DeepLabV3~\cite{deeplabv3} trained with different normalizations on PASCAL VOC 2012~\cite{pascal} validation set. Output stride is 16, without multi-scale or flipping when testing.}
    \label{tab:voc}
\end{table}

After evaluating BCN and WS on classification and detection, we test it on dense prediction tasks.
We start with semantic segmentation on PASCAL VOC~\cite{pascal}.
We choose DeepLabV3~\cite{deeplabv3} as the evaluation model for its good performances and its use of the pre-trained ResNet-101 backbone.

Table~\ref{tab:voc} shows our results on PASCAL VOC, which has $21$ different categories with background included.
We take the common practice to prepare the dataset, and the training set is augmented by
the annotations provided in \cite{pascalextra}, thus has 10,582 images.
We take our ResNet-101 pre-trained on ImageNet and finetune it for the task.
Here, we list all the implementation details for easy reproductions of our results:
the batch size is set to $16$, the image crop size is $513$, the learning rate follows polynomial decay with an initial rate $0.007$.
The model is trained for $30K$ iterations, and the multi-grid is $(1,1,1)$ instead of $(1, 2, 4)$.
For testing, the output stride is set to $16$, and we do not use multi-scale or horizontal flipping test augmentation.
As shown in Table~\ref{tab:voc}, by only changing the normalization methods from BN and GN to our BCN, mIoU increases by about $2\%$, which is a significant improvement for PASCAL VOC dataset.
As we strictly follow the hyper-parameters used in the previous work, there could be even more room of improvements if we tune them to favor BCN or WS, which we do not explore in this paper and leave to future work.

\subsection{Video Recognition on Something-Something}

\begin{table}[]
\small
\setlength{\tabcolsep}{0.6em}
    \centering
    \begin{tabular}{lc|cccc|cc}
        \toprule
        Model & \#Frame & GN & BN & WS & BCN & Top-1 & Top-5 \\
        \midrule
        RN50 & 8 & \cmark & & & & 42.07 & 73.20 \\
        RN50 & 8 & \cmark & & \cmark & & 44.26 & 75.51 \\
        \midrule
        RN50 & 8 & & \cmark & & & 44.30 & 74.53 \\
        RN50 & 8 & & \cmark & \cmark & & 46.49 & 76.46 \\
        \midrule
        RN50 & 8 & & & \cmark & \cmark & 45.27 & 75.22 \\
        \bottomrule
    \end{tabular}
    \caption{Comparing video recognition accuracy of TSM~\cite{tsm} on Something-SomethingV1~\cite{something}.}
    \label{tab:tsm}
\end{table}

In this subsection, we show the results of applying our method on video recognition on Something-SomethingV1 dataset~\cite{something}.
Something-SomethingV1 is a video dataset which includes a large number of video clips that show humans performing pre-defined basic actions.
The dataset has 86,017 clips for training and 11,522 clips for validation.

We use the state-of-the-art method TSM~\cite{tsm} for video recognition, which uses a ResNet-50 with BN as its backbone network.
The codes are based on TRN~\cite{trn} and then adapted to TSM.
The reimplementation is different from the original TSM~\cite{tsm}:
we use models pre-trained on ImageNet rather than Kinetics dataset~\cite{kinetics} as the starting points.
Then, we fine-tune the pre-trained models on Something-SomethingV1 for 45 epochs.
The batch size is set to 32 for 4 GPUs, and the learning rate is initially set to 0.0125, then divided by 10 at the 26th and the 36th epochs.
The batch normalization layers are not fixed during training.
With all the changes, the reimplemented TSM-BN achieves top-1/5 accuracy 44.30/74.53, higher than 43.4/73.2 originally reported in the paper.

Then, we compare the performances when different normalization methods are used in training TSM.
Table~\ref{tab:tsm} shows the top-1/5 accuracies of TSM when trained with GN, GN+WS, BN and BN+WS.
From the table we can see that WS increases the top-1 accuracy about $2\%$ for both GN and BN.
The improvements help GN to cache up the performances of BN, and boost BN to even better accuracies, which roughly match the performances of the ensemble TSM with 24 frames reported in the paper.
Despite that BCN improves performances of GN, it does not surpass BN.
This shows the limitation of BCN.

\section{Conclusion}
In this paper, we proposed two novel normalization methods, Weight Standardization (WS) and Batch-Channel Normalization (BCN) to bring the success factors of Batch Normalization (BN) into micro-batch training, including 1) the smoothing effects on the loss landscape and 2) the ability to avoid harmful elimination singularities along the training trajectory.
WS standardizes the weights in convolutional layers and BCN leverages estimated batch statistics of the activations in convolutional layers.
We provided theoretical analysis to show that WS reduces the Lipschitz constants of the loss and the gradients, and thus it smooths the loss landscape.
By investigating normalization methods from the perspective of elimination singularities, we found that channel-based normalization methods, such as Layer Normalization (LN) and Group Normalization (GN) are unable to keep far distances from elimination singularities, caused by lack of batch knowledge.
We showed that WS is able to alleviate this issue and BCN can further push models away from elimination singularities by incorporating estimated batch statistics channel-normalized models.
Experiments on comprehensive computer vision tasks, including image classification, object detection, instance segmentation, video recognition and semantic segmentation, demonstrate 1)  WS and BCN improve micro-batch training significantly, 2) WS+GN with batch size 1 is even able to match or outperform the performances of BN with large batch sizes, and 3) replacing GN by BCN leads to further improvement.

\ifCLASSOPTIONcaptionsoff
  \newpage
\fi



%



\bibliographystyle{IEEEtran}
\bibliography{egbib}

%

\begin{IEEEbiography}[{\includegraphics[width=1in,height=1.25in,clip,keepaspectratio]{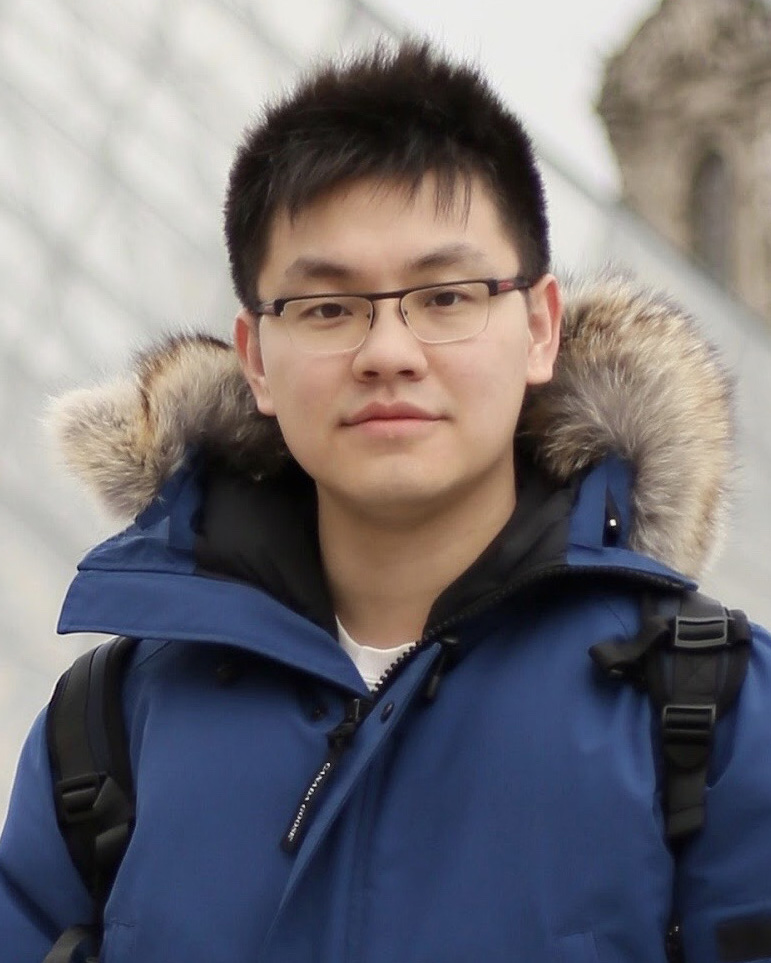}}]{Siyuan Qiao}
received B.E. in Computer Science at Shanghai Jiao Tong University in 2016.
He is currently a Ph.D. student at Johns Hopkins University, where he is advised by Bloomberg Distinguished Professor Alan Yuille.
From June 2017 to August 2017, he worked at Baidu IDL as an intern.
He interned at Adobe Inc. from June 2018 to August 2018.
He has also spent time at University of California, Los Angeles, and YITU Technology.
His research interests are computer vision and deep learning.
\end{IEEEbiography}

\begin{IEEEbiography}[{\includegraphics[width=1in,height=1.25in,clip,keepaspectratio]{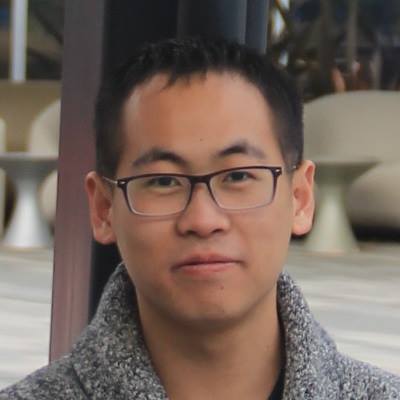}}]{Huiyu Wang}
is a Ph.D. student in Computer Science at Johns Hopkins University, advised by Bloomberg Distinguished Professor Alan Yuille.
He received M.S. in Electrical Engineering at University of California, Los Angeles in 2017 and B.S. in Information Engineering at Shanghai Jiao Tong University in 2015.
He also spent wonderful summers at Google, Allen Institute for Artificial Intelligence(AI2), and TuSimple.
His research interests are computer vision and machine learning.
\end{IEEEbiography}

\begin{IEEEbiography}[{\includegraphics[width=1in,height=1.25in,clip,keepaspectratio]{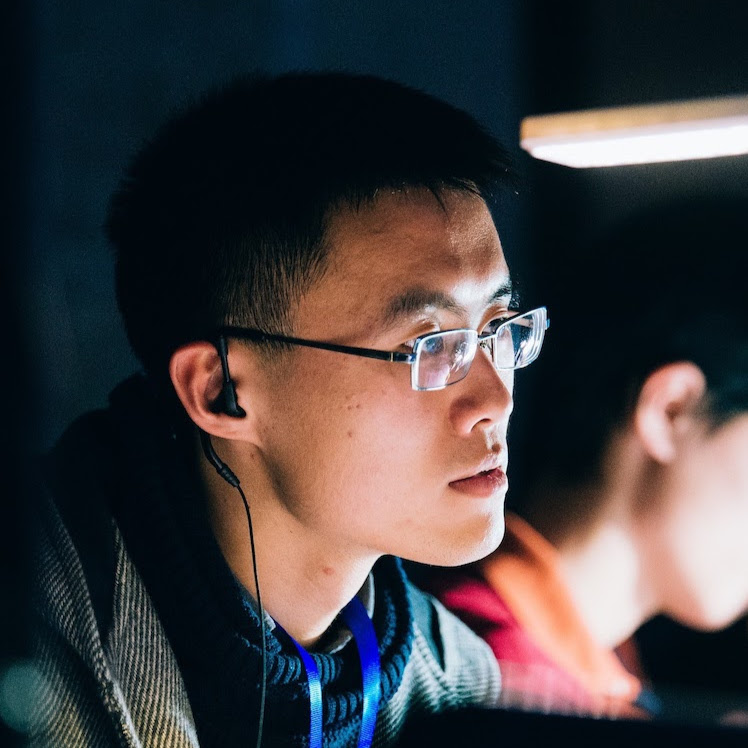}}]{Chenxi Liu}
is a Ph.D. student at Johns Hopkins University, where his advisor is Bloomberg Distinguished Professor Alan Yuille.
Before that, he received M.S. at University of California, Los Angeles and B.E. at Tsinghua University.
He has also spent time at Facebook, Google, Adobe, Toyota Technological Institute at Chicago, University of Toronto, and Rice University.
His research lies in computer vision and natural language processing.
\end{IEEEbiography}

\begin{IEEEbiography}[{\includegraphics[width=1in,height=1.25in,clip,keepaspectratio]{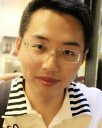}}]{Wei Shen}
received his B.S. and Ph.D. degree
both in Electronics and Information Engineering
from the Huazhong University of Science and
Technology, Wuhan, China, in 2007 and in 2012.
From April 2011 to November 2011, he worked
in Microsoft Research Asia as an intern. In 2012,
he joined the School of Communication and Information Engineering, Shanghai University and
served as an assistant and associate professor
until Oct 2018. He is currently an Assistant Research Professor at the Department of Computer
Science, Johns Hopkins University. His current research interests include computer vision, deep learning and biomedical image analysis.
\end{IEEEbiography}

\begin{IEEEbiography}[{\includegraphics[width=1in,height=1.25in,clip,keepaspectratio]{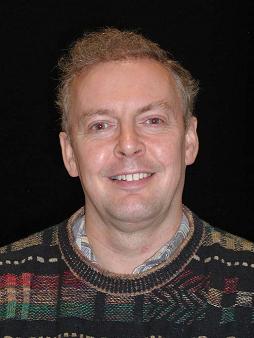}}]{Alan Yuille}
received his B.A. in mathematics
from the University of Cambridge in 1976, and
completed his Ph.D. in theoretical physics at
Cambridge in 1980. He then held a postdoctoral
position with the Physics Department, University
of Texas at Austin, and the Institute for Theoretical Physics, Santa Barbara. He then became
a research scientists at the Artificial Intelligence
Laboratory at MIT (1982-1986) and followed this
with a faculty position in the Division of Applied
Sciences at Harvard (1986-1995), rising to the
position of associate professor. From 1995-2002 he worked as a senior
scientist at the Smith-Kettlewell Eye Research Institute in San Francisco.
From 2002-2016 he was a full professor in the Department of Statistics
at UCLA with joint appointments in Psychology, Computer Science, and
Psychiatry. In 2016 he became a Bloomberg Distinguished Professor in
Cognitive Science and Computer Science at Johns Hopkins University.
He has won a Marr prize, a Helmholtz prize, and is a Fellow of IEEE.
\end{IEEEbiography}

\end{document}